\documentclass[journal]{IEEEtran}

\usepackage{latexsym}
\usepackage[ruled,linesnumbered,noresetcount,vlined]{algorithm2e}

\SetKwFor{Repeat}{repeat}{:}{endw}
\usepackage{mdframed}
\usepackage{xspace}
\usepackage[export]{adjustbox}

\usepackage{authblk}
\usepackage{float}
\floatstyle{ruled}
\newfloat{model}{thp}{lop}
\floatname{model}{Model}

\usepackage{multicol}

\usepackage{amsmath,amsfonts,mathrsfs,amssymb,bm,mathtools,mathabx}
\usepackage{times, helvet, courier}
\usepackage{graphicx, epsfig, epstopdf}
\usepackage{wrapfig, multirow, multicol}
\usepackage{rotating, booktabs,longtable}
\usepackage{url}
\usepackage{stmaryrd,xspace,arydshln}
\usepackage{picinpar}
\usepackage{todonotes}
\usepackage{color}
\usepackage{hyperref}
\hypersetup{
    colorlinks=true,
    linkcolor=cyan,
    citecolor=purple,
    urlcolor=blue,
    filecolor=blue
}

\def\thmspace{0.2em}
\newtheorem{theorem}{\hspace{\thmspace}{\bf Theorem}\!}
\newtheorem{corollary}{\hspace{\thmspace}{\bf Corollary}\!}
\newtheorem{definition}{\hspace{\thmspace}{\bf Definition}\!}
\newtheorem{property}{\hspace{\thmspace}{\bf Property}\!}

\newenvironment{proof}{{\textit{Proof}.}}{\hfill$\Box$}

\def \iff{\;\Leftrightarrow\;}

\newcommand{\benumerate}{\begin{list}{$\bullet$}{\topsep=0pt \parsep=0pt \itemsep=1pt \labelwidth=1.5em \labelsep=0.5em \leftmargin=20pt}}
\newcommand{\eenumerate}{\end{list}}
\newcommand{\bitemize}{\begin{list}{$\bullet$}{\topsep=0pt \parsep=0pt \itemsep=1pt \leftmargin=10pt}}
\newcommand{\eitemize}{\end{list}}

\newcommand{\setf}[1]{{\bf{#1}}}

\newcommand{\cA}{\mathcal{A}}
\newcommand{\cD}{\mathcal{D}}

\newcommand{\cO}{{\mathcal{O}}}	%
\newcommand{\cN}{\mathcal{N}}

\newcommand{\cR}{\mathcal{R}}

\newcommand{\bY}{\setf{Y}}

\newcommand{\bx}{\setf{x}}

\newcommand{\bg}{\setf{g}}

\newcommand{\bb}{\setf{b}}
\newcommand{\bS}{\setf{S}}
\newcommand{\bs}{\setf{s}}
\newcommand{\bv}{\setf{v}}

\newcommand{\RR}{\mathbb{R}}

\newcommand{\Lap}{{\text{Lap}}}

\begin{document}
\title{Differential Privacy for Power Grid Obfuscation}

 \author{
   \IEEEauthorblockN{
                                  }
 }

\author{Ferdinando~Fioretto,
        Terrence~W.K.~Mak,~\IEEEmembership{Member,~IEEE,} 
        and~Pascal~Van~Hentenryck,~\IEEEmembership{Member,~IEEE}
\thanks{The authors are affiliated with the 
H. Milton Stewart School of Industrial and Systems Engineering,
Georgia Institute of Technology, Atlanta, GA 30332, USA
e-mail: fioretto@gatech.edu, wmak@gatech.edu, pvh@isye.gatech.edu.}
}

\markboth{IEEE Transactions on Smart Grid}%
{}

\maketitle\sloppy\allowdisplaybreaks

\begin{abstract}
  The availability of high-fidelity energy networks brings significant value to academic and commercial research. However, such releases also raise fundamental concerns related to privacy and security as they can reveal sensitive commercial information and expose system vulnerabilities. This paper investigates how to release the data for power networks where \emph{the parameters of transmission lines and transformers are obfuscated}. It does so by using the framework of Differential Privacy (DP), that provides strong privacy guarantees and has attracted significant attention in recent years. Unfortunately, simple DP mechanisms often result in AC-infeasible networks. To address these concerns, this paper presents a novel differentially private mechanism that guarantees AC-feasibility and largely preserves the fidelity of the obfuscated power network. Experimental results also show that the obfuscation significantly reduces the potential damage of an attack carried by exploiting the released dataset. 
  \end{abstract}

\IEEEpeerreviewmaketitle

\section{Introduction}

The availability of test cases representing high-fidelity power system networks is fundamental to foster research in optimal power flow, unit commitment, and transmission planning, to name only a few challenging problems. This need was recognized by ARPA-E when it initiated the \emph{Grid Data Program} in 2015. However, the release of such rich datasets is challenging due to legal issues related to privacy and national security. For instance, the electrical load of an industrial customer indirectly exposes sensitive information on its production levels and strategic investments, and the value parameters of lines and generators may reveal how transmission operators operate their networks. Furthermore, this data could be exploited by an attacker to inflict targeted damages on the network infrastructure.

This paper explores whether differential privacy can help to mitigate these concerns. Differential Privacy (DP) \cite{Dwork:06} is an algorithmic property that measures and bounds the privacy risks associated with answering sensitive queries or releasing a privacy-preserving dataset. It introduces carefully calibrated noise to the data to prevent the disclosure of sensitive information. An algorithm satisfying DP offers privacy protection regardless of the external knowledge of an attacker. In particular, the definition of DP adopted in this paper ensures that an attacker obtaining access to a differentially private output, cannot detect (with high probability)  how close is the privacy-preserving value to its original one.

However, DP faces significant challenges when the resulting privacy-preserving datasets are used as inputs to complex optimization algorithms, e.g., \emph{Optimal Power Flow} (OPF) problems. Indeed, the privacy-preserving dataset may have lost the fidelity and realism of the original data and may even not admit feasible solutions for the optimization problems of interest \cite{fioretto:CPAIOR-18}.

This paper studies how to address such a challenge when the goal is to
preserve the privacy of line parameters and transformers. It presents
a DP mechanism to release data for power networks that is realistic
and limits the power of an attacker. More precisely, the contribution
of this paper is fourfold. {\it (1)} It proposes the \emph{Power Line
  Obfuscation} (PLO) mechanism to obfuscate the line parameters in a
power system network.  {\it (2)} It shows that PLO has strong
theoretical properties: It achieves $\epsilon$-differential privacy,
it ensures that the released data can produce feasible solutions for
OPF problems, and its objective value is a constant factor away from
optimality.  {\it (3)} It extends the PLO mechanism to handle
time-series network data.  {\it (4)} It demonstrates, experimentally,
that the PLO mechanism improves the accuracy of existing
approaches. When tested on the largest collection of OPF test cases
available, it results in solutions with similar costs and optimality
gaps to those obtained on the original problems, while also protecting
well against an attacker that has access to the released data and uses
it to damage the real power network. Interestingly, \emph{on the test
  cases adopted, the damage inflicted on the real power network, when
  the attacker exploits the PLO-obfuscated data, converges to that of
  a random, uninformed attack as the privacy level increases.}


\section{Related Work}

There is a rich literature on theoretical results of DP (see, for
instance, the excellent surveys \cite{Dwork:13} and \cite{Vadhan:17}).
The literature on DP applied to energy systems includes considerably
fewer efforts.  {\'A}cs and Castelluccia \cite{Acs:11} exploit a
direct application of the Laplace mechanism to hide user participation
in smart meter data sets, achieving $\epsilon$-DP.  Zhao et
al.~\cite{Zhao:14} study a DP schema that exploits the ability of
households to charge/discharge a battery to hide the real energy
consumption of their appliances.  Halder et al.~\cite{halder2017}
propose an DP-based architecture to support privacy-preserving
thermal inertial load management at an aggregated level.  Liao et
al.~\cite{Liao:17} introduce Di-PriDA, a privacy-preserving mechanism
for appliance-level peak-time load balancing control in the smart
grid, aimed at masking the consumption of the top-k appliances of a
household.

Karapetyan et al.~\cite{Karapetyan:17} empirically quantify the
trade-off between privacy and utility in demand response systems. The
authors analyze the effects of a simple Laplace mechanism on the
objective value of the demand response optimization problem. Their
experiments on a 4-bus micro-grid show drastic results: the optimality
gap approaches nearly 90\% in some cases. Eibl and Engel \cite{eibl2017}
studied the effect of DP-based aggregation schemes on the utility of
real smart meter data, and \cite{eibl2018} studies the impact of
applying a DP approach to protect metering data used for load
forecasting.  Zhou et al.~\cite{zhoudifferential} have recently
studied the problem of releasing differential private network
statistics obtained from solving a \emph{Direct Current} (DC) optimal
power flow problem.

A differential private schema that uses constrained post-processing
was recently introduced by Fioretto et al.~\cite{fioretto:CPAIOR-18}
and adopted to protect load consumption in power networks. In
contrast, the proposed PLO mechanism releases the obfuscated network
data protecting the line parameters imposing constraints to ensure
that the problem solution cost is close to the solution cost of the
original problem, and that the underlying optimal power flow
constraints are satisfiable.


\newcommand{\refacpf}{(\hyperref[model:ac_pf]{AC-OPF})}

\section{Preliminaries}
\label{sec:preliminaries}

This section reviews the AC Optimal Power Flow (AC-OPF) problem and
key concepts from differential privacy.  A summary of the notation
adopted is tabulated in Table \ref{tab:notation}. Bold-faced symbols
are used to denote constant values.

\begin{table*}[!t]
        \caption{Power Network Nomenclature.\label{tab:notation}}
\begin{center}
      \resizebox{0.9\linewidth}{!}
      {
      \begin{tabular}{l l  l l}
        \toprule
        {$N$}               & The set of nodes in the network& 
        {$\theta^\Delta$}   & Phase angle difference limits\\
        {$E$}               & The set of {\em from} edges in the network&  
        {$S^d = p^d+ \bm iq^d$} & AC power demand\\
        {$E^R$}             & The set of {\em to} edges in the network&  
        {$S^g = p^g+ \bm iq^g$} & AC power generation\\
        {$\bm i$}           & Imaginary number constant&
        {$c_0,c_1,c_2$}     & Generation cost coefficients\\
        $n$, $m$               & $|E|$ and $|N|$, respectively &
        {$\Re(\cdot)$}      & Real component of a complex number\\
        {$S = p+ \bm iq$}   & AC power&
        {$\Im(\cdot)$}      & Imaginary component of a complex number\\
        {$V = v \angle \theta$}  & AC voltage&
        {$Y = g + \bm ib$}  & Line admittance\\
        {$|\cdot|$}         & Magnitude of a complex number&
        {$\angle$}          & Angle of a complex number \\
        {$s^u$}             & Line apparent power thermal limit&
        {$x^l$, $x^u$}      & Lower and upper bounds of $x$\\
        {$\theta_{ij}$}     & Phase angle difference (i.e., $\theta_i - \theta_j$)&
        {$\bm x$} &         A constant value \\
        {$\cN$}   &         A network description &
        {$\bg$}   &         Network's line conductances\\
        {$\bb$}   &         Network's line susceptances&
        {$\Delta_Q$}   &    Query sensitivity\\
        {$\epsilon$}   &    Privacy budget&
        {$\alpha$}   &      Indistinguishability value\\
        {$\beta$}   &       Faithfulness parameter&
        {$\tilde{\bx}$} &   Privacy-preserving version of $\bx$\\
        {$\dot{\bx}$} &     Post-processed version of $\tilde{\bx}$&
        {${\bx}^{*}$} &     Complex conjugate of ${\bx}$\\
         \bottomrule
        \end{tabular}
        }
    \end{center}

\end{table*}

\subsection{AC Optimal Power Flow}
\label{sec:notation}

\emph{Optimal Power Flow (OPF)} is the problem of determining the most economical generator dispatch to serve demands while satisfying
operating and feasibility constraints. \emph{AC-OPF}
refers to modeling the full AC power equations when computing an OPF.
This paper views the grid as a graph $(N,E)$ where $N$ is the set of
buses and $E$ is the set of transmission lines and transformers,
called \emph{lines} for simplicity. We use $E$ to represent the set of directed arcs and $E^R$ to refer to the arcs in $E$ with the reverse direction. The AC power flow equations use complex quantities
for current $I$, voltage $V$, admittance $Y$, and power $S$. The
quantities are linked by constraints expressing Kirchhoff's Current
Law (KCL)
and Ohm's Law,
resulting in the AC Power Flow equations:
%
\begin{align*}
  & S^g_i - {\bm S^d_i} = \sum_{\substack{(i,j)\in E \cup E^R}} S_{ij} \;\; \forall i\in N \\ 
  & S_{ij} = \bm Y^*_{ij} |V_i|^2 - \bm Y^*_{ij} V_i V^*_j \;\; (i,j)\in E \cup E^R  
\end{align*}

\noindent These non-convex nonlinear equations are the core building
block in many power system applications and Model~\ref{model:ac_opf}
depicts the AC-OPF formulation.  The objective function \eqref{ac_obj}
minimizes the cost of the generator dispatch.  Constraint
\eqref{eq:ac_0} sets the reference angle for a slack bus $s \in N$ to
be zero to eliminate numerical symmetries.  Constraints
\eqref{eq:ac_1} and \eqref{eq:ac_6} capture the voltage bounds and
phase angle difference constraints.  Constraints \eqref{eq:ac_2} and
\eqref{eq:ac_5} enforce the generator output and line flow limits.
Finally, Constraints \eqref{eq:ac_3} and \eqref{eq:ac_4} capture the AC Power Flow equations.  We use $\cN = \langle N, E, \bS, \bY,
\bm{\theta^\Delta}, \bs, \bv \rangle$ for a succinct \emph{network
  description} and define $m=|N|$ and $n = |E|$.

\begin{model}[!t]
  \caption{The AC Optimal Power Flow Problem (AC-OPF)}
  \label{model:ac_opf}
  \vspace{-6pt}
  {\small
  \begin{align}
    \mbox{\bf variables:} \;\;
    & S^g_i, V_i \;\; \forall i\in N, \;\;
      S_{ij}   \;\; \forall(i,j)\in E \cup E^R \nonumber \\
    \mbox{\bf minimize:} \;\;
    & \sum_{i \in N} \bm c_{2i} (\Re(S^g_i))^2 + \bm c_{1i}\Re(S^g_i) + \bm c_{0i} \label{ac_obj} \\
    \mbox{\bf subject to:} \;\; 
    & \angle V_{s} = 0,  \label{eq:ac_0} \\
    & \bm {v^l}_i \leq |V_i| \leq \bm {v^u}_i     \;\; \forall i \in N \label{eq:ac_1} \\
    & -\bm {\theta^\Delta}_{ij} \leq \angle (V_i V^*_j) \leq \bm {\theta^\Delta}_{ij} \;\; \forall (i,j) \in E  \label{eq:ac_6}  \\
    & \bm {S^{gl}}_i \leq S^g_i \leq \bm {S^{gu}}_i \;\; \forall i \in N \label{eq:ac_2}  \\
    & |S_{ij}| \leq \bm {s^u}_{ij}          \;\; \forall (i,j) \in E \cup E^R \label{eq:ac_5}  \\
    & S^g_i - {\bm S^d_i} = \textstyle\sum_{(i,j)\in E \cup E^R} S_{ij} \;\; \forall i\in N \label{eq:ac_3}  \\ 
    & S_{ij} = \bm Y^*_{ij} |V_i|^2 - \bm Y^*_{ij} V_i V^*_j       \;\; \forall (i,j)\in E \cup E^R \label{eq:ac_4}
  \end{align}
  }
  \vspace{-12pt}
\end{model}


\subsection{Differential Privacy}
\label{sec:differential_privacy}

\emph{Differential privacy} \cite{Dwork:06} is a privacy framework that protects the disclosure of the participation of an individual to a dataset. In the context of this paper, differential privacy is used to protect the disclosure of conductance and susceptance values of transmission lines.  The paper considers datasets $D = \{ g_1, \ldots, g_n \} \in \RR^n$ as $n$-dimensional real-valued vector describing conductance values and aims at protecting the value of each conductance $g_i$ up to some quantity $\alpha > 0$. This requirement allows to \emph{obfuscate} the parameters of lines whose values are close to one another while keeping the distinction between line parameters that are far apart from each other. This privacy notion is characterized by the \emph{adjacency relation} between datasets that captures the \emph{indistinguishability} of individual line values and defined as: 
\begin{equation} \label{eq:adj_rel} 
    D \sim_\alpha D' \iff \exists i
    \textrm{~s.t.~} | g_i - g'_i | \leq \alpha \textrm{~and~} g_j = g'_j,
    \forall j \neq i.  
\end{equation} 
where $D$ and $D'$ are two datasets and $\alpha > 0$ is a real
value \cite{chatzikokolakis2013broadening}. Informally speaking, a DP
mechanism needs to ensure that queries on two indistinguishable datasets (i.e., datasets differing on a single value by at most $\alpha$)
give similar results. The following definition formalizes this
intuition \cite{Dwork:06,chatzikokolakis2013broadening}.

\begin{definition}
A randomized algorithm $\cA \!:\! \cD \!\to\! \cR$ with domain $\cD$ and range $\cR$ is $\epsilon$-differential private if, for any output response $O \subseteq \cR$ and any two \emph{adjacent} inputs $D \sim_\alpha D' \in \RR^n$, fixed a value $\alpha > 0$, 
\begin{equation}
  \label{eq:dp_def} 
  \frac{Pr[\cA(D) \in O]}{Pr[\cA(D') \in O]} \leq \exp(\epsilon).
\end{equation}
\end{definition}

\noindent 

The level of \emph{privacy} is controlled by the parameter $\epsilon
\geq 0$, called the \emph{privacy budget}, with small values denoting
strong privacy. The level of \emph{indistinguishability} is controlled
by the parameter $\alpha > 0$. Differential privacy satisfies several
important properties, including \emph{composability} and
\emph{immunity to post-processing} \cite{Dwork:13}.

\begin{theorem}[Sequential Composition]
\label{th:seq_composition}
The composition $(\cA_1(D), \ldots, \cA_k(D))$ of a collection
$\{\cA_i\}_{i=1}^k$ of $\epsilon_i$-differential private algorithms
satisfies $(\sum_{i=1}^{k}
\epsilon_i)$-differential privacy.
\end{theorem}

\begin{theorem}[Parallel Composition] 
\label{th:par_composition} 
Let $D_1$ and $D_2$ be disjoint subsets of $D$ and $\cA$ be an
$\epsilon$-differential private algorithm.  Computing $\cA(D
\cap D_1)$ and $\cA(D \cap D_2)$ satisfies $\epsilon$-differential privacy.
\end{theorem}

\begin{theorem}[Post-Processing Immunity] 
\label{th:postprocessing} 
Let $\cA$ be an $\epsilon$-differential private algorithm and $g$ be
an arbitrary mapping from the set of possible output sequences to an
arbitrary set. Then, $g \circ \cA$ is $\epsilon$-differential private.
\end{theorem}

A function (also called \emph{query}) $Q: \RR^n \to \RR$ can be made
differential private by injecting random noise to its output. The
amount of noise to inject depends on the \emph{sensitivity} of the
query, denoted by $\Delta_Q$ and defined as
\[
\Delta_Q = \max_{D \sim_\alpha D'} \left\| Q(D) - Q(D')\right\|_1.
\]
For instance, querying the conductance values of a line from a dataset
$D$ is achieved through an identity query $Q$, whose sensitivity
$\Delta_Q = \alpha$. The Laplace distribution with 0 mean and scale
$b$, denoted by $\Lap(\lambda)$, has a probability density function
$\Lap(x|\lambda) = \frac{1}{2\lambda} e^{-\frac{|x|}{\lambda}}$. It
can be used to obtain an $\epsilon$-differential private algorithm to
answer numeric queries \cite{Dwork:06}.  In the following,
$\Lap(\lambda)^n$ denotes the i.i.d.~Laplace distribution over $n$
dimensions with parameter $\lambda$.

\begin{theorem}[Laplace Mechanism]
\label{th:m_lap} 
Let $Q$ be a numeric query that maps datasets to $\RR^n$. The Laplace
mechanism that outputs $Q(D) + z$, where $z \in \RR^n$ is drawn from
the Laplace distribution
$\textrm{\Lap}\left(\frac{\Delta_Q}{\epsilon}\right)^n$, achieves
$\epsilon$-differential privacy.
\end{theorem}

\section{The Obfuscation Mechanism for Power Lines}
\label{sec:PGO}

\subsection{Problem Setting and Attack Model}
\newcommand{\cev}[1]{\reflectbox{\ensuremath{\vec{\reflectbox{\ensuremath{#1}}}}}}

A power grid operator desires to release a network description
$\tilde{\cN} \!=\! \langle N, E, \bS, \tilde{\bY}, \bm{\theta^\Delta},
\bs, \bv \rangle$ of a network $\cN \!=\! \langle N, E, \bS, \bY,
\bm{\theta^\Delta}, \bs, \bv \rangle$, that \emph{obfuscates} the
lines admittance values $\bY$ within a given indistinguishability
parameter $\alpha$. In addition, the released data must preserve the realism of the original network. The lines parameters are considered to be extremely sensitive as they can reveal important operational information that can be exploited by an attacker to inflict targeted damages on the network infrastructure~\cite{Schainker2006,McDaniel2009}.  The paper assumes that the optimal dispatch cost $\cO(\cN)$ and typical conductance-susceptance line ratios are publicly available and thus accessible by an attacker. This is not restrictive since the optimal dispatch cost can be inferred from market clearing prices which are publicly accessible, and the ratios between conductance and susceptance of a line can be retrieved from the manufacture informational material.


This paper also considers an attack model in which a malicious user can disrupt $k$ power lines (called \emph{attack budget}) to inflict maximal network damage. It further assumes that the attacker has full knowledge of a network description $\cN$ and can use it to estimate the damages inflicted by its actions.  To measure the impact of an attack on the power network, this paper measures the total load affected and the amount of the affected loads that can be restored via Model~\ref{model:load_res}. The latter aims at maximizing the active load $l_i \Re(S^d_i)$ served by the damaged network while preserving the active/reactive factor.

\begin{model}[!t]
  \caption{Maximum Load Restoration}
  \label{model:load_res}
  \vspace{-6pt}
  {\small
  \begin{align}
    \mbox{\bf variables:} \;\;
    & S^g_i, V_i, l_i \;\; \forall i\in N, \;\;
      S_{ij}   \;\; \forall(i,j)\in E \cup E^R \nonumber \\
    \mbox{\bf maximize:} \;\;
    & \sum_{i \in N}  l_i \Re(S^d_i) \label{ac_obj2} \\
    \mbox{\bf subject to:} \;\; 
    & (\ref{eq:ac_0})\text{--}(\ref{eq:ac_4})   \\
    & 0 \leq l_i  \leq 1 \;\; \forall i\in N \label{eq:load_res_1}  \\
    & S^g_i - l_i {\bm S^d_i} = \textstyle\sum_{(i,j)\in E \cup E^R} S_{ij} \;\; \forall i\in N \label{eq:load_res_2}  
  \end{align}
  }
  \vspace{-12pt}
\end{model}

\subsection{The PLO Problem}

The \emph{Power Lines Obfuscation} (PLO) problem establishes the fundamental desiderata to be delivered by the obfuscation mechanism. It operates on the line conductances and suceptances, which are denoted by $\bg =\{ g_{ij}\}_{(i,j)\in E}$ and $\bb =\{ b_{ij}\}_{(i,j)\in E}$, respectively.

\begin{definition}[PLO problem]
	\label{def:plop}
	Given a network description $\cN$ and positive real values
        $\alpha, \beta,$ and $\epsilon$, the PLO problem produces a
        network description $\tilde{\cN}$ that satisfies:

	\begin{enumerate}

	\item \label{tag:cond1}
		\emph{Lines obfuscation}: The lines conductances
          $\tilde{\bg}$ of $\tilde{\cN}$ satisfy
          $\epsilon$-differential privacy under
          $\alpha$-indistinguishability.

	\item \label{tag:cond2}
		\emph{Consistency}: $\tilde{\cN}$ have feasible solutions to the OPF Constraints \eqref{eq:ac_0}--\eqref{eq:ac_4}.

	\item \label{tag:cond3}
		\emph{Objective faithfulness}: $\tilde{\cN}$ must be faithful to
          the value of the objective function up to a factor $\beta$,
          i.e., $\frac{| \cO(\cN) - \cO(\tilde{\cN}) |}{\cO(\cN)} \leq
          \beta $.
	\end{enumerate}

\end{definition}

\noindent Finding values $\tilde{\bg}$ satisfying
$\alpha$-indistinguishability is readily achieved through the Laplace
mechanism. However, finding values $\tilde{\bY}$ that satisfy
conditions (2) and (3) is more challenging. Indeed, these conditions
require that the new network $\tilde{\cN}$ satisfies the AC power flow
equations, the operational constraints, and closely preserves the
objective value.  In other words, these conditions ensure the realism
and fidelity of $\tilde{\cN}$.

\subsection{The PLO Mechanism}

The \emph{PLO} mechanism, described in Algorithm \ref{alg:plo}, addresses these challenges. It takes as input a power network description $\cN$, its optimal dispatch cost, denoted as $\cO^*$, as well as three positive real numbers: $\epsilon$, which determines the \emph{privacy value} of the private data, $\alpha$, which determines the \emph{indistinguishability value}, and $\beta$, which determines the required \emph{faithfulness} to the value of the objective function. 

The PLO mechanism relies on three, independent, private data estimations, each of which requires the addition of carefully calibrated noise to guarantee privacy. 
It first injects Laplace noise with parameter $\lambda=3\alpha/\epsilon$ to the output of each query on each dimension of the conductance vector $\bg$ of $\cN$ resulting in new noisy conductance and susceptances vectors:
\begin{equation}
\label{eq:ID_query}  
    \tilde{\bg} = \bg + Lap\left(\frac{3\alpha}{\epsilon}\right)^n,     \qquad     \tilde{\bb} = \textbf{r} \cdot \tilde{\bg}, 
\end{equation}
as shown in lines \ref{ln:1}--\ref{ln:2} of Algorithm \ref{alg:plo},
where $\tilde{\bg} = \{\tilde{g}_{ij}\}_{(ij)\in E}$ and $\tilde{\bb}
= \{\tilde{b}_{ij}\}_{(ij)\in E}$ are the vectors of noisy
conductances and suceptances, $\textbf{r}$ is the vector of ratios
$\{\frac{g_{ij}}{b_{ij}}\}_{(ij)\in E}$ between $\bg$ and $\bb$, and
$\cdot$ denotes the dot-product. Note that, importantly, the mechanism
retains the conductance-susceptance ratio.

It is also important to ensure that the line values within different voltage levels preserve their differences. Denote by $\textsl{VL}(\cN)$ as the set of voltage levels in $\cN$. For each voltage level $v$, the PLO mechanism computes the noisy mean value of the conductance vector $\bg$ (lines \ref{ln:3} and \ref{ln:4}):
\begin{equation}
\label{eq:AVG_query_B}
	\tilde{\mu}_{\bg}^v = \left(     
	\frac{1}{n_v} \sum_{(ij) \in E(v)} g_{ij} \right)     +
	\text{Lap}\left(\frac{3\alpha}{n_v\epsilon} \right), 
\end{equation}
 and
suceptance $\bb$: 
\begin{equation}
\label{eq:AVG_query_B1}
\tilde{\mu}_\bb^v = 
	\left(\frac{1}{n_v} \sum_{(ij) \in E(v)} b_{ij} \right)  +
	\text{Lap}\left(\frac{3\alpha}{n_v\epsilon} \right), 
\end{equation}
where $E(v)$ denotes the subset of lines at voltage level $v$, and $n_v = |E(v)|$.  These estimates are used to guarantee that the parameters of the lines do not deviate too much from their original values within each voltage level.

While the application of the Laplace noise to produce new conductance
and susceptance vectors satisfies condition \ref{tag:cond1} of
Definition \ref{def:plop}, it may not satisfy conditions
\ref{tag:cond2} and \ref{tag:cond3}. In fact, Section \ref{sec:exp}
shows that the Laplace noise induced on the parameters of the lines
often result in a new network description which admits no feasible
flow. To overcome this limitation, the PLO mechanism
\emph{post-processes} the noisy values $\tilde{\bb}$ and $\tilde{\bg}$
by exploiting an optimization model specified in line 5 of Algorithm
\ref{alg:plo}. The result of such an optimization-based
post-processing step is a new network $\dot{\cN} = \langle N, E, \bS,
\dot{\bY}, \bm{\theta^\Delta}, \bs, \bv \rangle$ that satisfies the
objective faithfulness and consistency requirements of
Definition~\ref{def:plop}.

The optimization model minimizes the sum of the $L_2$-distances
between the variables $\dot{\bg} \in \RR^n$ and the noisy conductances
$\tilde{\bg} \in \RR^n$, and the variables $\dot{\bb} \in \RR^n$ and
the noisy susceptances $\tilde{\bb} \in \RR^n$. The model is subject
to Constraints \eqref{eq:ac_0}--\eqref{eq:ac_3} of Model
\ref{model:ac_opf} with the addition of the $\beta$-faithfulness
constraint \eqref{dp_beta_con} that guarantees to satisfy condition
\ref{tag:cond3} of the PLO problem (Definition \ref{def:plop}). The notation $\bm{cost}(S^g_i)$ is used as a shorthand for $c_{2i} (\Re(S^g_i))^2 + \bm 
c_{1i}\Re(S^g_i) + \bm c_{0i}$.  
Constraint \eqref{dp_ac_4} enforces
the power-flow based on the Ohm's Law on the post-processed
conductance and suceptance values. Finally, Constraints
\eqref{dp_bound_g} and \eqref{dp_bound_b} bound the values for the
post-processed conductance and susceptance to be close to their \emph
{privacy-preserving} means, parameterized by a value $\lambda > 0$.  These constraints are used to avoid that the post-processed values deviate
arbitrarily from the original ones. The optimization also works on a
pre-processed network to ensure the realism and feasibility. This
pre-processing ensures that parallel lines have the same resistance
and reactance parameters and that negative resistance values are not
part of the obfuscation. Moreover, the optimization guarantees that
all remaining resistances are positive.

\let\oldnl\nl
\newcommand{\nonl}{\renewcommand{\nl}{\let\nl\oldnl}}

\begin{algorithm}[!t]
	\caption{The PLO mechanism for the AC-OPF}
	\label{alg:plo}
	\SetKwInOut{Input}{input}
    \SetKwInOut{Output}{output}	
	\Input{$\langle \cN, \cO^*, \epsilon, \alpha, \beta \rangle$}

	$\tilde{\bg} \gets \bg + \text{Lap}(\frac{3\alpha}{\epsilon})^n$ 
	\label{ln:1}
	\\
	$\tilde{\bb} \gets \textbf{r} \cdot \tilde{\bg} $ 
	\label{ln:2}
	\\
	\ForEach{$v \in \textsl{VL}(\cN)$}{
	\label{ln:3}
		$\tilde{\mu}_{\bx}^v \gets \frac{1}{n} \sum_{(ij)\in E(v)} x_{ij} 
		+ \text{Lap}(\frac{3\alpha}{n_v\epsilon})$ (for $x = {\bg, 
		\bb}$)
		\label{ln:4}
	}
	Solve the following model:
	\label{ln:5}
	\nonl
	{\small
	\begin{align}
		\mbox{\bf variables: }
		& S^g_i, V_i            \hspace{123pt} \forall i \in N \notag \\
		& \dot{Y}_{ij}, S_{ij}  \hspace{77pt} \forall (i,j)\in E_k \!\cup\! E_k^R \notag\\
		\mbox{\bf minimize: }  
		&  \| \dot{\bg} - \tilde{\bg} \|_2^2  + \| \dot{\bb} - \tilde{\bb} \|_2^2   \tag{$s_1$}\label{dp_obj}\\
		\mbox{\bf subject to:} & \notag \\
		& \hspace*{-20pt}  	\eqref{eq:ac_0}\text{--}\eqref{eq:ac_3} \notag \\ 
		& \hspace*{-20pt} 	\frac{\big| \sum_{i \in N} \bm{cost}(S_i^g) - \cO^* \big|}{\cO^*} \leq \beta 
		\tag{$s_2$} \label{dp_beta_con} \\
		& \hspace*{-20pt} 
	    S_{ij} = \dot{Y}^*_{ij} |V_i|^2 - \dot{Y}^*_{ij} V_i V^*_j  \;\; \forall (i,j)\in E \cup E^R 		\tag{$s_3$} \label{dp_ac_4} \\
		& \hspace*{-20pt} 
		\forall (i,j)\in E(v) \cup E^R(v),\, \forall v \in \textsl{VL}(\cN): \notag\\ 
		& \hspace*{-0pt}
		\frac{1}{\lambda}\mu_{\bg}^v \leq \dot{g}_{ij} \leq \lambda\,\mu_{\bg}^v \;\;  \tag{$s_4$} \label{dp_bound_g} \\
		& \hspace*{-0pt}
		\frac{1}{\lambda}\mu_{\bb}^v \leq \dot{b}_{ij} \leq \lambda\,\mu_{\bb}^v \;\; \tag{$s_5$} \label{dp_bound_b}
	\end{align}\\
	\Output{$\dot{\cN} = \langle  N, E, \bS, \dot{\bY}, \bm{\theta^\Delta}, \bs, \bv \rangle$}
	}
\end{algorithm}

The PLO mechanism can be thought as redistributing the noise of the
Laplace mechanism applied to the admittance values of the lines in
$\cN$ to obtain a new network $\dot{\cN}$ that is consistent with the
problem constraints and objective. It searches for a feasible solution
that satisfies the AC-OPF constraints and the $\beta$-faithfulness
constraint.

The following results show that PLO has desirable properties.  All the
proofs are in Appendix \ref{sec:missing-proofs}.

\begin{theorem}
For a given $\alpha$-indistinguishability level, the PLO mechanism is $\epsilon$-differential private.
\end{theorem}

The following result is a consequence of \cite{fioretto:CPAIOR-18}(Theorem 5).

\begin{corollary}
\label{th:opt}
The optimal solution $\langle \bg^*, \bb^*\rangle$ to the optimization model in 
line 5 of Algorithm \ref{alg:plo}
 satisfies
$
\| \bg^* - \bg \|_{2} + \| \bb^* - \bb \|_{2} 
\leq 
2 \| \tilde{\bg} - \bg \|_{2} + 2 \| \tilde{\bb} - \bb \|_{2}.
$
\end{corollary}

\noindent

This last result implies the PLO mechanism is at most a factor 2 away
from optimality. Such a result, in turn, follows from the optimality
of the Laplace mechanism for identity queries
\cite{koufogiannis:15}. Note that a solution to the PLO always
exists. Indeed, the original network values $\bm Y$ represent a
feasible solution satisfying all requirements of Definition
\ref{def:plop}.


\section{Extension: Multi-step PLO}
\label{sec:mplo}

The PLO mechanism obfuscates the line parameters based on a single snapshot (time-point) of the steady state of the network. 
To improve the fidelity and realism of the network, the mechanism can be generalized by reasoning about multiple snapshots (time-series)
of loads and optimal dispatches.
Consider a set of network descriptions $\{ \cN_t \}_{t=1}^h$ over a finite time horizon $h$ where the loads and the optimal generator dispatches are varying over time. Each $\cN_t = (N, E, \bS_t, \bY, \bm{\theta^\Delta}, \bs, \bv)$ represents the steady state of the power grid at time step $0 < t \leq h$, and therefore $\{ \cN_t \}_{t=1}^h$ represents a network of \emph{time-series} data.

The \emph{Multi-step Power Lines Obfuscation} (MPLO) problem extends the PLO by enforcing the \emph{objective faithfulness} condition (Definition \ref{def:plop}) for every $\cN_t$ ($t \in [h]$).  The derived MPLO mechanism is outlined in Appendix \ref{sec:MPLO-alg}. It extends the PLO mechanism in taking as input a collection of networks $\{\cN_t\}_{t=1}^h$ and it returns the admittance values $\dot{\bY}$ for the lines in the power grid that satisfy the conditions of Definition \ref{def:plop}.  
The MPLO mechanism differs from the PLO mechanism exclusively in the post-processing optimization step of line 5. 
Observe that the admittance matrix $\bY$ containing the line parameters is constant during the whole time horizon. Therefore, similarly to PLO, the MPLO mechanism \emph{perturbs the lines parameters only once}, prior applying the post-processing optimization step. Thus, the MPLO mechanism provides the same privacy guarantees as those provided by the PLO mechanism. 
\begin{theorem}\label{thm:mplo}
For a given $\alpha$-indistinguishability level, the MPLO mechanism is $\epsilon$-differential private.
\end{theorem}

The objective of MPLO is equivalent to the objective \eqref{dp_obj} of
the post-processing step in the PLO mechanism.  In addition, the model
constraints naturally extend those of the PLO mechanism by considering
multiple time steps with a fixed time horizon.  The MPLO mechanism
also outputs a new network whose line parameters $\dot{\bY}$ are
\emph{obfuscated} and do not deviate too far from their
privacy-preserving mean values.  However, it further requires the
AC-OPF problem constraints and the $\beta$ consistency requirement are
satisfied for the whole time series instead of just one time-step.
The PLO mechanism can be seen as a special case of the MPLO mechanism
with a time horizon $h = 1$; Hence the PLO mechanism is a relaxation
of MPLO mechanism.


\section{Experimental Evaluations}
\label{sec:exp}

This section examines the proposed mechanisms on a variety of networks
from the NESTA library \cite{Coffrin14Nesta}. It analyzes the line
values produced by the obfuscation procedure, studies the mechanism
ability to preserve the dispatch costs and optimality gaps (to be
defined shortly), determines how well the resulting network can
sustain an attack, and reports the runtime of the mechanisms. It also
extends this analysis to time-series networks using a multiple-step
approach.

For presentation simplicity, the analysis focuses primarily on the
IEEE 39-bus network. However, our results are consistent across the
entire NESTA benchmark set.  All experiments use a privacy budget
$\epsilon = 1.0$ and vary the \emph{indistinguishability level}
$\alpha \in \{10^{-3}, 10^{-2}, 10^{-1}, 1.0\}$ and the
\emph{faithfulness level} $\beta \in \{10^{-2}, 10^{-1}\}$.  The model
was implemented using the Julia package
PowerModels.jl~\cite{Coffrin:18} with the nonlinear solver
IPOPT~\cite{wachter06on} for solving the various power flow models,
including the nonlinear AC model and the
QC~\cite{coffrin16the,hijazi13convex} and SOCP~\cite{jabr06radial}
relaxations.

\subsection{Analysis of the Line Parameters}

\begin{figure*}[!t]
	\centering
	\includegraphics[width=0.99\linewidth]{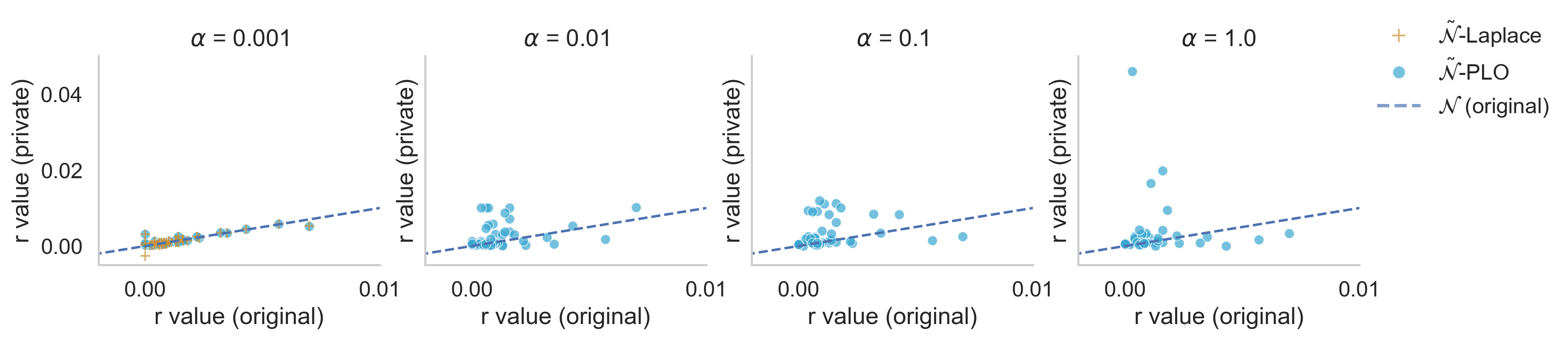}
	\caption{IEEE-39 bus line resistances (p.u.) at varying of the indistinguishability level $\alpha \in \{0.001, 0.01, 0.1, 1.0\}$ and for $\beta=0.01$. The x-axis shows values of the original network and the y-axis shows values of the PLO and Laplace obfuscated networks.} 
	\label{fig:r_plo1}
\end{figure*}

This section studies the realism of the Laplace
mechanism. Table~\ref{tbl:lap-feasibility} reports the percentage of
feasible instances (over 100 runs) for obfuscated networks obtained
using \emph{exclusively} the Laplace mechanism on the IEEE-30 bus,
IEEE-39 bus, IEEE-57 bus, and the IEEE-118 bus networks.  When the
indistinguishability values $\alpha$ exceed $0.1$, the
Laplace-obfuscated networks are rarely AC-feasible. In contrast, the
PLO mechanism is \emph{always} AC-feasible (except for one IEEE-118
instance).  These results justify the need of studying mechanisms that
are more sophisticated than the Laplace mechanism, and hence the PLO
mechanism.  Figure~\ref{fig:r_plo1} illustrates the line resistances
of an IEEE 39-bus network obtained by the Laplace and the PLO
mechanisms and compare them with the associated values in the original
network. The figure reports the results at varying of the
indistinguishability level $\alpha$ and fixing $\beta=0.01$.
\emph{The results indicate that the OPF obfuscated values differ by at
most 1\% from their original ones.}  Not surprisingly, the differences
are more pronounced as the indistinguishability level increases: For
larger indistinguishability levels, the PLO mechanism introduces more
noise and hence more diverse lines values are generated.


 \begin{table}[!t]
	\centering
	\caption{Laplace mechanism feasibility (\%)}
	\resizebox{0.75\linewidth}{!}
	{
	\begin{tabular}{lrrr}
	\toprule
	& \multicolumn{3}{c}{$\alpha$}\\
	\cmidrule(lr){2-4} 
	Network instance &   {0.001} & {0.01} & $\geq${0.1} \\
	\midrule
	nesta\_case30\_ieee      &   100 & 80 & 0  \\
	nesta\_case39\_epri      &    100 & 0 & 0  \\
	nesta\_case57\_ieee      &    100 & 61 & 0 \\
	nesta\_case118\_ieee     &    100 & 47 & 0 \\
	\bottomrule
\end{tabular}
}
\label{tbl:lap-feasibility} 
\end{table}

\subsection{Dispatch Costs and Optimality Gaps}

\begin{figure}[!t]
	\centering
	\includegraphics[width=\linewidth]{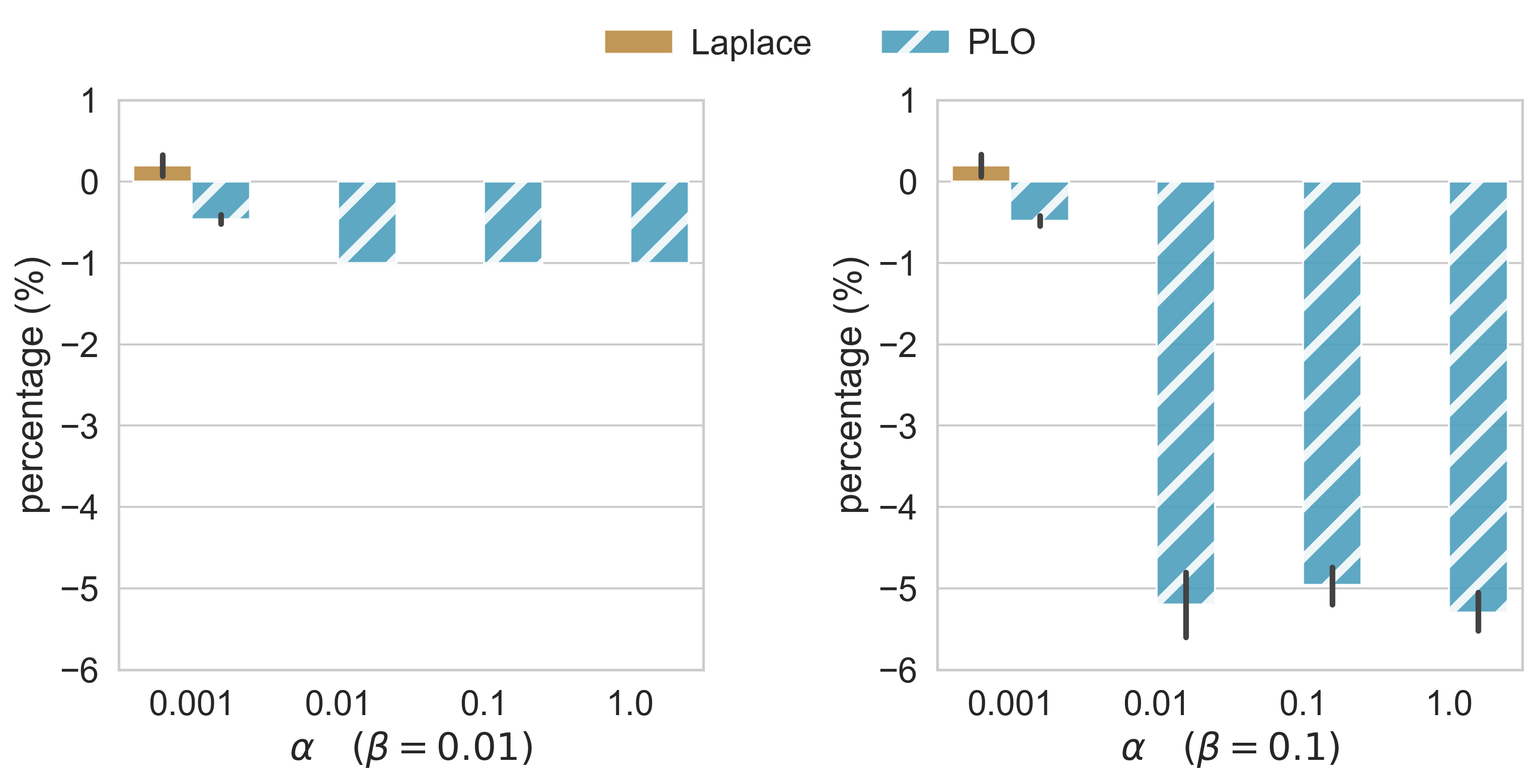}
	\caption{The IEEE-39 bus AC optimal dispatch costs differences, in percentage, between the original and the obfuscated networks with faithfulness parameters $\beta=0.01$ (left) and $0.1$ (right).}
	\label{fig:opf_12a}
\end{figure}

The next results evaluate the ability of the PLO-obfuscated networks
to preserve the dispatch costs and optimality gaps.  Optimality gaps
measures the relative distance between the objective value of the
AC-OPF problem with one of its relaxations.  It is frequently used as
a measure to reflect the hardness of a problem instance.  For our
experimental evaluations, it is used as a proxy to measure whether the
structure of an instance changed after applying our proposed
mechanism.
Figure~\ref{fig:opf_12a} shows the difference, in percentage, of the
dispatch costs obtained via the Laplace and the PLO mechanisms with
respect to the original costs at varying of the indistinguishability
level $\alpha$ and the faithfulness parameter $\beta$.  The figure
illustrates the mean and the standard deviation (shown with black,
solid, lines) obtained on $100$ runs, for each combination of the
$\alpha$ and $\beta$ parameters. Optimality gap (percentage
differences) are measured as: $100 \times \frac{\cO(\cN) -
\cO(\tilde{\cN}) }{ \cO(\cN)}$, where $\cN$ represents the original
network, $\tilde{\cN}$ the obfuscated network (using the Laplace or
the PLO mechanisms), and $\cO$ the cost of a (local) optimal solution
to the AC-OPF problem. Parameter $\alpha$ controls the amount of noise
being added to the line parameters, therefore, the OPF costs are close
to their original values when $\alpha$ is small (e.g., $\leq
10^{-3}$). The PLO mechanism often produces obfuscated network
inducing OPFs with lower costs than the original ones. This is because
PLO returns an AC-feasible solution whose cost is close to that of the
original network, ignoring whether a lower dispatch cost may exist.

 \begin{figure}[!t]
	\centering
	\includegraphics[width=\linewidth]{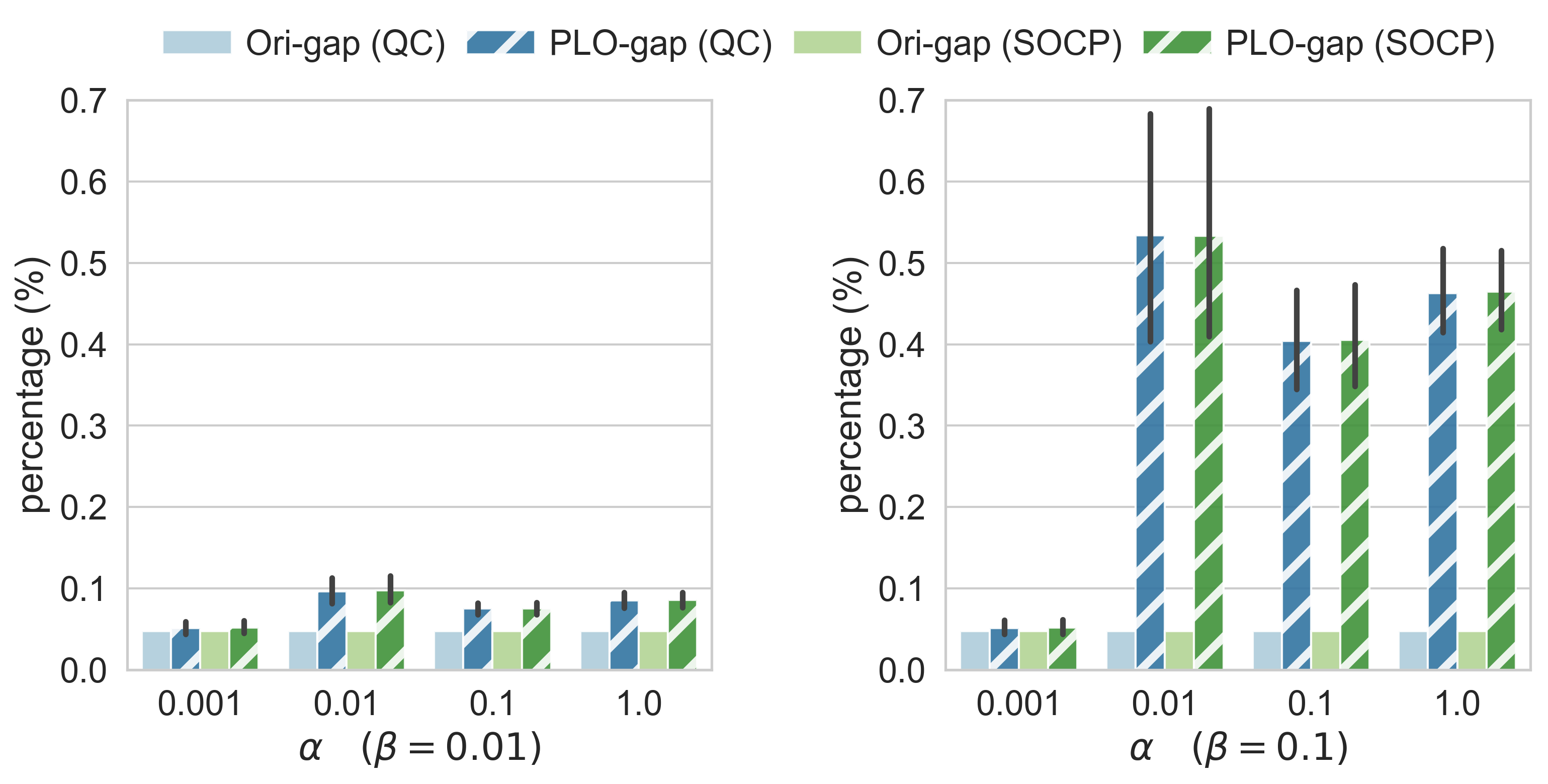}
	\caption{The IEEE-39 bus optimality gap differences, in percentage, between the original and the obfuscated networks with faithfulness parameters $\beta=0.01$ (left) and $0.1$ (right).}
	\label{fig:opf_12b}
\end{figure}

Figure~\ref{fig:opf_12b} compares the optimality gaps on the QC and
the SOCP relaxations of the AC-OPF obtained using the original and the
PLO-obfuscated networks.  The percentage measures are defined as $100
\times \frac{|\cO(\cN) - \widehat{\cO}(\cN)|}{\cO(\cN)}$, where $\cN$
is either the original or the obfuscated network, and $\widehat{\cO}$
is the function returning the costs of the QC or the SOCP AC-OPF
relaxation on $\cN$.  The results are averaged on $100$ runs and show
that the optimality gaps attained with the obfuscated networks are
close to those attained with the original ones for small ($\leq
10^{-2}$) $\alpha$ values. \emph{This is important for capturing the
fidelity of the obfuscated network and the difficulty of the
associated OPF.} In general, the PLO mechanism increases the
optimality gaps slightly, and these results are consistent across the
NESTA networks.

\subsection{Power Grid Attack Simulation}

The next experiment evaluates the damages that an attacker may inflict
on a real power network $\cN$, if an obfuscated counterpart
$\tilde{\cN}$ is released. The attack setting is as follows: An
attacker is given a budget denoting the percentage $k$ of lines it can
damage. The attacker chooses the lines to damage based on the
obfuscated network, but the attack impact is evaluated on the real
network. To assess the benefits of the proposed obfuscation scheme, in
response to such an attack, three attack strategies are compared:

\begin{enumerate} 
	\item \label{random_attack}
	\emph{Random Attack}: $k$ lines are randomly selected. This represents a scenario in which an attacker carries an \emph{uninformed} attack.

	\item \label{obfuscated_flow_attack}
	\emph{Obfuscated Flow Attack}: 
	The attacker solves an OPF problem on the \emph{obfuscated network} $\tilde{\cN}$ and chooses the top-$k$ lines carrying the largest active flows. This case represents a scenario in which the attacker carries an informed attack based on the obfuscated network data.

	\item \label{real_flow_attack}
	\emph{Real Flow Attack}: The attacker solves an OPF problem on
	  the \emph{real network} $\cN$ and chooses the top-$k$ lines carrying the largest active flows. This case represents a scenario in which the \emph{real data} is released and exploited by an attacker.
\end{enumerate}

\noindent

\begin{figure}[!t]
	\centering
	\includegraphics[width=\linewidth]{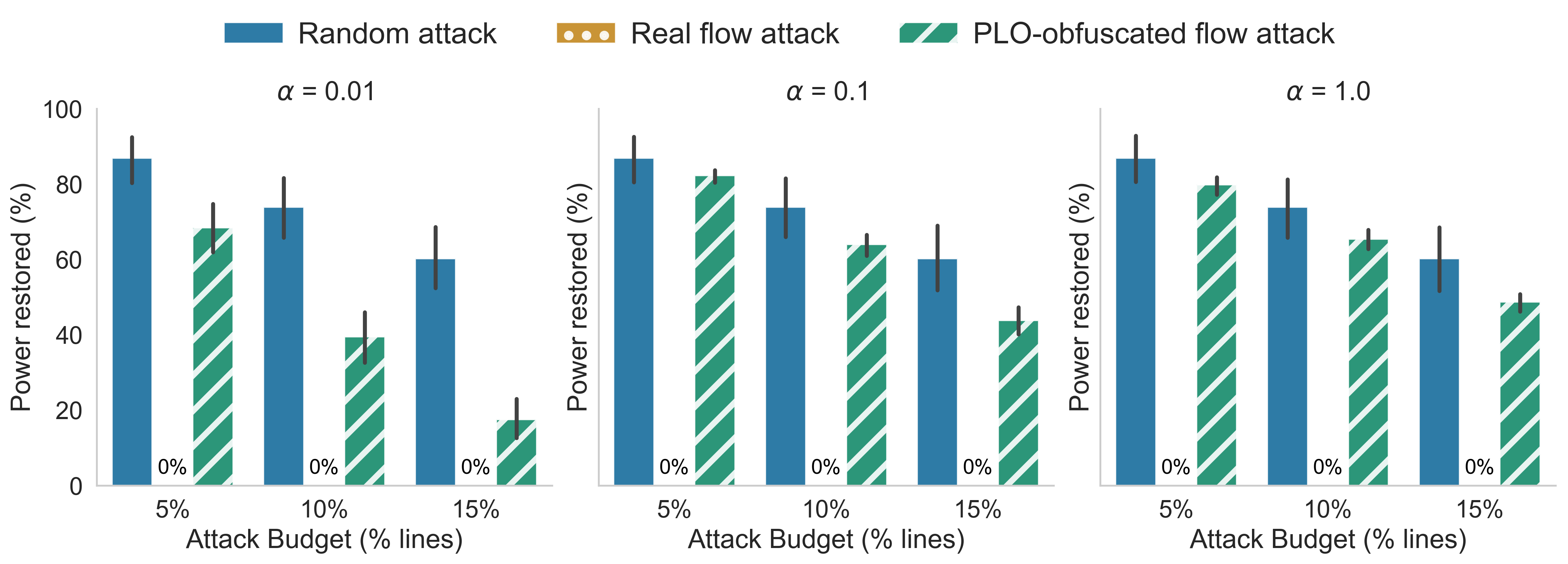}

	\caption{Percentage of the active load restored after different attacks on the IEEE-39 bus network bus network with $\alpha=0.01$ (left), $0.1$ (center), and $1.0$ (right), and $\beta=0.01$.}
	\label{fig:attack_1}
\end{figure}

\begin{figure}[!t]
	\centering
	\includegraphics[width=\linewidth]{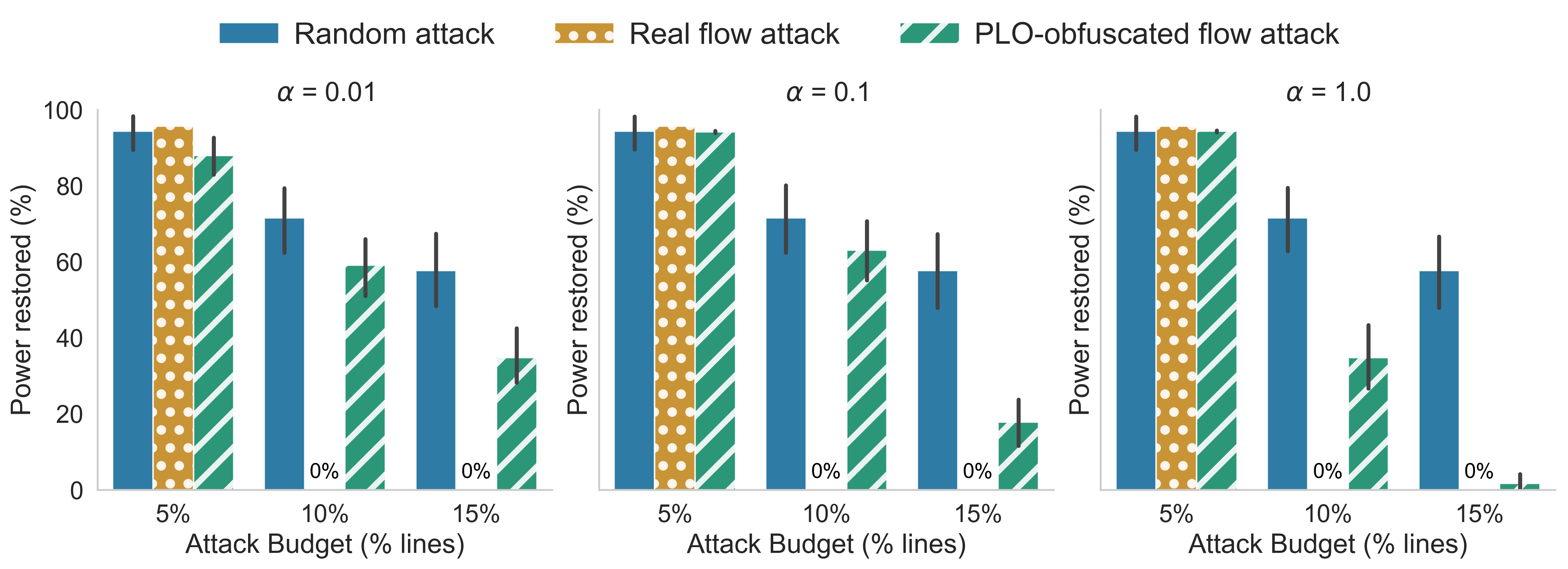}

	\caption{Percentage of the active load restored after different attacks on the IEEE-118 bus network with $\alpha=0.01$ (left), $0.1$ (center), and $1.0$ (right), and $\beta=0.01$.}
	\label{fig:attack_2}
\end{figure}

To compare the damages inflicted by the attacks, the experiments
report the load that can be restored after an attack using the
\emph{maximum load restoration model} described in Model
\ref{model:load_res}. Figures~\ref{fig:attack_1} and
\ref{fig:attack_2} illustrate the percentage of the load being
restored for each attack strategies, at varying of the attack budget
$k \in \{5, 10, 15\}$ and the indistinguishability value $\alpha \in
\{0.01, 0.1, 1.0\}$, on the IEEE-39 bus and the IEEE-118 bus
benchmarks, respectively, with faithfulness value $\beta = 0.01$.  The
results report the average values of 100 simulations for each
combination of parameters.

The \emph{random attacks} are used as a baseline to assess the damage that may be caused by an uninformed attacker. Not surprisingly, they result in the largest load restoration for each setting and the restored load decreases as the attack budget increases. In contrast, the \emph{real flow attacks} produce the most significant damage to the networks. In all cases tested, the load restoration after these attacks were close to $0$\% after the attacker reaches a sufficient budget (as low as 5\% for IEEE-39 bus and 10\% for IEEE-118 bus),
meaning that \emph{these attacks are highly effective and extremely harmful}. 

On the other hand, the results for the \emph{Obfuscated Flow Attacks} show a different pattern. Even though the attacker selects the lines with the highest flows (in the obfuscated network), the network ability to restore loads is substantially higher when compared to those of the real flow attacks. \emph{Remarkably, as the indistinguishability values increase, the strength of the obfuscated flow attack to inflict damages decreases and its success rate are close to those of random attacks}.  This is because larger indistinguishability implies more noise and thus higher chance for an attacker to damage less harmful lines.

\subsection{Mechanism Runtimes}

Having shown the effectiveness of the PLO mechanism in generating
obfuscated networks, we now analyze its computational efficiency.
Table~\ref{tbl:comp1} tabulates the average runtime, in seconds, for
100 experiments on several NESTA instances at varying of the
indistinguishability values ($\alpha$ in $ \{ 0.01, 0.1, 1.0\}$) and
using the faithfulness value $\beta=0.01$.  \emph{In all cases,
producing an obfuscated network requires less than $60$ seconds.} The
results illustrate that, in general, the runtime increases when
$\alpha$ increases.  Larger $\alpha$ values may result in obfuscated
line parameters that are farther from the original values, thus
affecting the power losses and the feasible power flows. Therefore,
minimizing the PLO optimization model (line \ref{ln:5} of Algorithm
\ref{alg:plo}) may increase the runtime.

 \begin{table}[!t]
	\centering
	\small
	\caption{PLO Computational Runtime}
	\resizebox{0.8\linewidth}{!}
	{
	\begin{tabular}{lrrr}
	\toprule
	& \multicolumn{3}{c}{$\alpha$}\\
	\cmidrule(lr){2-4} 
	Network instance &  {0.01} & {0.1} & {1.0} \\
	\midrule
	nesta\_case3\_lmbd       &      0.04 &      0.05 &      0.07 \\
	nesta\_case4\_gs         &      0.09 &      0.14 &      0.16 \\
	nesta\_case5\_pjm        &      0.10 &      0.09 &      0.15 \\
	nesta\_case6\_c          &      0.05 &      0.12 &      0.19 \\
	nesta\_case6\_ww         &      0.05 &      0.26 &      0.39 \\
	nesta\_case9\_wscc       &      0.08 &      0.21 &      0.29 \\
	nesta\_case14\_ieee      &      0.10 &      0.44 &      0.74 \\
	nesta\_case24\_ieee\_rts  &      0.63 &      1.04 &       1.88 \\
	nesta\_case29\_edin      &      4.23 &      3.26 &      4.59 \\
	nesta\_case30\_as        &      0.38 &      1.31 &      1.77 \\
	nesta\_case30\_fsr       &      0.39 &      1.43 &      1.70 \\
	nesta\_case30\_ieee      &      0.43 &      1.41 &      1.63 \\
	nesta\_case39\_epri      &      1.67 &      2.00 &      2.25 \\
	nesta\_case57\_ieee      &      1.11 &      3.43 &      4.81 \\
	nesta\_case73\_ieee\_rts  &      2.86 &      8.03 &      13.90 \\
	nesta\_case89\_pegase    &     33.67 &      44.96 &       48.12 \\
	nesta\_case118\_ieee     &      8.79 &      8.64 &      17.73 \\
	nesta\_case162\_ieee\_dtc &     17.48 &      38.83 &       50.43 \\
	\bottomrule
\end{tabular}
}
\label{tbl:comp1} 
\end{table}

\subsection{MPLO for Time-Series Data}

This subsection evaluates the effect of obfuscating a network with the
Multistep PLO mechanism.  The experiments use time-series data
$\{\cN_t\}_{t=1}^h$ with a time horizon of $h$ steps, obtained and
computed by varying the load profile in the range of
$[80\%, 110\%]$ of their original values. 

Each time step will be associated with a load profile.  The goal of
MPLO is to find line parameters that meet the constraints imposed by a
fixed number $1 \leq r \leq h$ time steps ($r = 1$ reduces to PLO, and
$r = h$ implies using all the steps).  These $r$ time steps are
equally spaced in the time horizon $[h]$.

\begin{figure}[!t]
	\centering
	\includegraphics[width=0.95\linewidth]{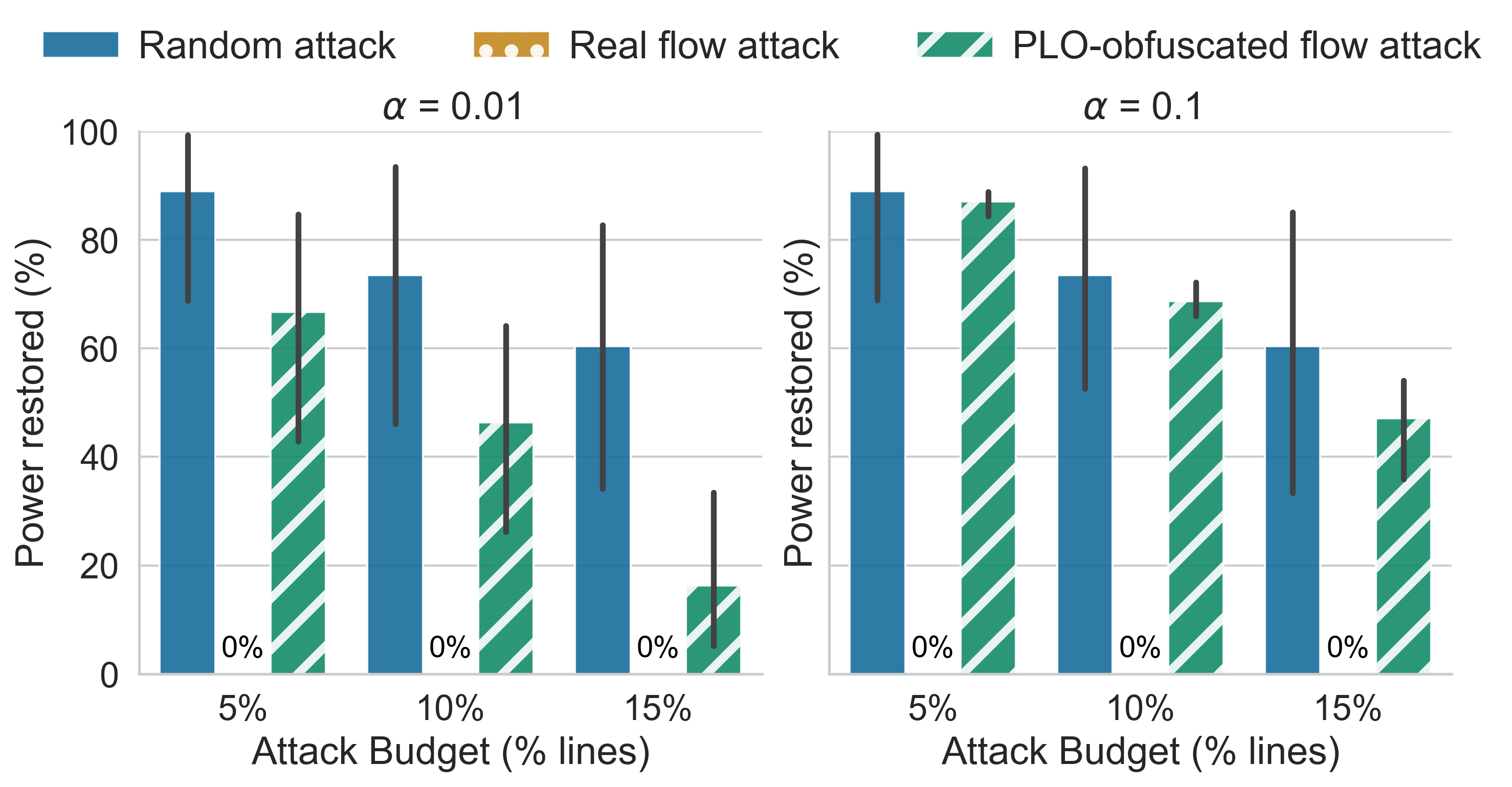}

	\caption{Active load restored (in percentage) after different attacks on the IEEE-39 bus network with $\alpha \in \{0.01, 0.1\}$, $\beta = 0.01$, and number $r=4$ of time step evaluated.}
	\label{fig:attack_2a}
\end{figure}

\begin{figure}[!t]
	\centering	
	\includegraphics[width=0.95\linewidth]{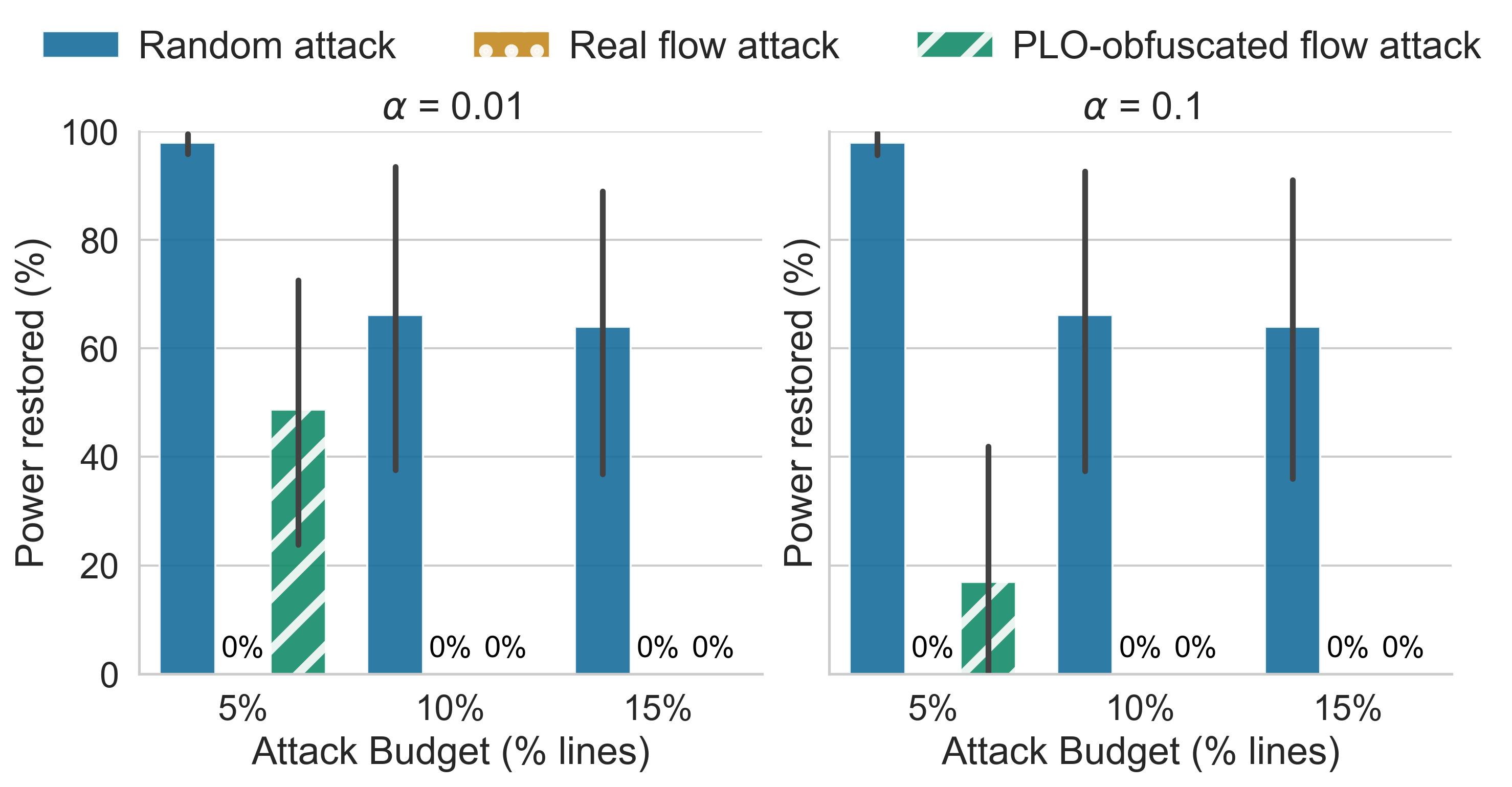}

	\caption{Active load restored (in percentage) after different attacks on the IEEE-39 bus network with $\alpha \in \{0.01, 0.1\}$, $\beta = 0.01$, and number $r=31$ of time step evaluated.}
	\label{fig:attack_2b}
\end{figure}

The results on the OPFs and optimality gaps are very similar to those
obtained by the (single-step) PLO mechanism.  The presentation focuses
on the power network attacks which have significant differences.
Figures~\ref{fig:attack_2a} and \ref{fig:attack_2b} illustrates the
percentage of the load restored on the IEEE-39 bus network for the
attack strategies and settings discussed earlier, using $r=4$ and
$r=h=31$, respectively. The results for $r=4$ are similar to those
obtained by the PLO mechanism. In contrast, when the entire time
horizon is considered, the effectiveness of attacks on the obfuscated
network increases. This is because constraining the MPLO optimization
problem to consider each single time step in the time horizon reduces
the degrees of freedom for generating obfuscated networks that differ
substantially from the original ones.

\begin{figure}[!t]
	\centering
	\includegraphics[width=0.6\linewidth]{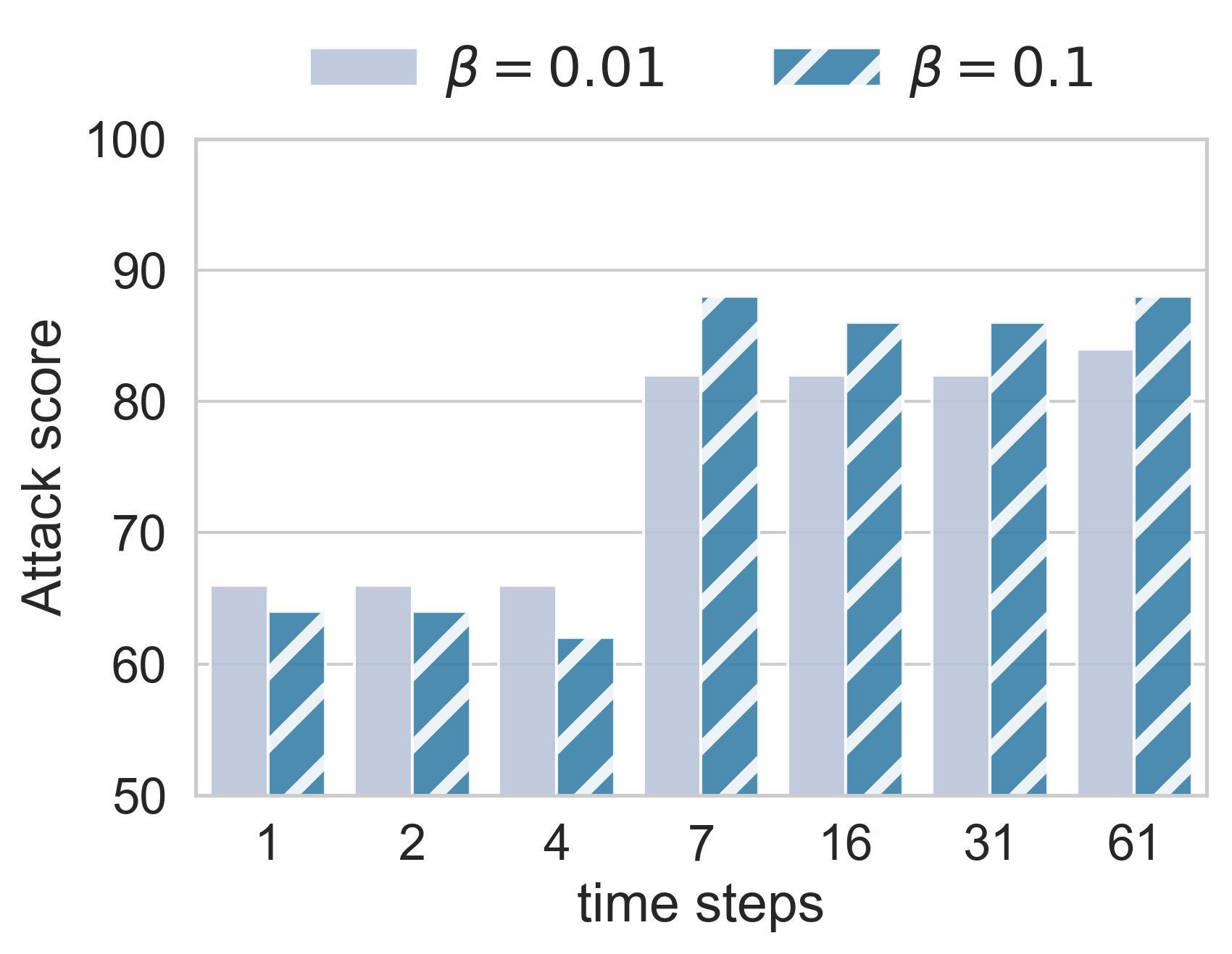}

	\caption{IEEE-39 bus' Attack score over an increasing number of time steps, with $\alpha \in \{0.01, 0.1\}$, $\beta=0.01$, and $k=10\%$.}
	\label{fig:attack_3}
\end{figure}

Figure~\ref{fig:attack_3} further explains these results.  It
quantifies the similarity of the attacks on the real and obfuscated
networks using the following metric to measure similarity: $100 \times
\frac{\lvert E^{o} \cap E^{d} \rvert}{ \lvert E^{o} \rvert}$ where
$E^o$ and $E^d$ are the set of lines selected by the real and
obfuscated flow attacks. The experiments fix the attack budget at $k=10\%$
and vary the number of time steps $r$ within the MPLO mechanism.
The figure clearly illustrates that the number of lines chosen by both attacks
grows as $r$ increases. When $r\geq 7$, the two attack configurations
select, on average, up to $90\%$
of common lines.  The experiment highlights the tradeoffs between
obfuscation and network fidelity: Larger values for $r$ result in
higher network fidelity but reduce the effects of the obfuscation
process, making the networks more vulnerable to attacks.

\section{Conclusions}

This paper presented a privacy-preserving scheme for the release of
power grid benchmarks that obfuscate the parameters of transmission
lines and transformers. The proposed Power Line Obfuscation (PLO)
mechanism hides the network sensitive values using differential
privacy, while also ensuring that the released obfuscated network
preserves fundamental properties useful in optimal power
flow. Specifically, the released networks have dispatch costs similar
to those of the original networks and satisfy the power flow
operational constraints.  The PLO mechanism was tested on a large
collection of test cases. It was shown to be efficient and to produce
obfuscated networks that preserve dispatch costs and optimality gaps
values. Finally, the networks released by the PLO mechanism are shown
to be effective in deceiving an attacker attempting to damage the
network components for disrupting the power grid load.  Future work
will focus on jointly obfuscating other sensitive aspects of the
network, such as loads and generators.  Another avenue of future
research is to study more complex attack models, including those in
which the attacker evaluates the (near)-optimal subset of lines to
disrupt so to minimize the total load restoration.

\section*{Acknowledgment}

The authors would like to thank Kory Hedman for extensive discussions
about obfuscation techniques and effective attack strategies. The
authors are also grateful to the anonymous reviewers for their
valuable comments.  This research is partly funded by the ARPA-E Grid
Data Program under Grant 1357-1530.

\bibliographystyle{IEEEtran}
\bibliography{differential_privacy}

\newpage
\appendices

\newtheorem{innercustomthm}{{\bf Theorem}}
\newenvironment{customthm}[1]
  {\renewcommand\theinnercustomthm{#1}\innercustomthm}
  {\endinnercustomthm}

  \section{Detailed Proofs}
  \label{sec:missing-proofs}

  This section provides the missing proofs. It first reviews the
  sensitivity of the queries adopted in the PLO mechanisms. The
  characterization of properties discussed below holds true for the
  $\alpha$-indistinguishability model of differential privacy
  \cite{chatzikokolakis2013broadening}.

\begin{property}
\label{prop:sens_id}
	Let $D = \{x_1, \ldots, x_n\} \in \RR^n$ be an $n$-dimensional numerical vector. 
	The sensitivity of the identity query $Q_I(D) = \{x_1, \ldots, x_n\}$ is $\Delta_{Q_I} = \alpha$. 
\end{property}
The property above follows directly from the definition of sensitivity
of queries in the $\alpha$-indistinguishability model.

\begin{property}
\label{prop:sens_avg}
	Let $D = \{x_1, \ldots, x_n\} \in \RR^n$ be an $n$-dimensional numerical vector. 
	The sensitivity of the average query $Q_A(D) = \frac{1}{n} \sum_{i=1}^n x_i$ 
	is $Q_A = \frac{\alpha}{n}$.
\end{property}
\begin{proof}
Let $D$ and $D'$ be two datasets such that $D \sim_\alpha D'$, that is $D'$ differs from $D$ in at most one coordinate and for a factor of at most $\alpha$. Denote by $i$ be the coordinate such that $|x_i - x_i'| \leq \alpha$. 
It follows:
\begin{align*}
	| Q_A(D) - Q_A(D') | &= 
	\left| \frac{1}{n} \sum_{j=1}^n x_j - \frac{1}{n} \sum_{j=1}^n x_j' \right|\\
	&= \frac{1}{n} |x_i - x_i' | \tag{by Eq.~\eqref{eq:adj_rel}}\\
	&\leq \frac{1}{n} \alpha.
\end{align*}
\end{proof}

\begin{customthm}{4}
	\label{thm4}
	For a given $\alpha$-indistinguishability level, the PLO mechanism is $\epsilon$-differential private.
\end{customthm}

\begin{proof}
	Consider an indistinguishability value $\alpha > 0$. 
	Algorithm \ref{alg:plo} queries the dataset of real conductance data in three different instances:
	\begin{enumerate}
		\item To compute $\tilde{\bg}$: 
		In Equation \eqref{eq:ID_query}, $\tilde{\bg}$ is computed by issuing an identity query over $\bg$. The privacy budget used in Equation \eqref{eq:ID_query} is $\frac{\epsilon}{3 \alpha}$. Thus, since by Property \ref{prop:sens_id}, Theorem \ref{th:m_lap}, and parallel composition (Theorem \ref{th:par_composition}), privately computing the conductance values $\tilde{\bg}$ is $\epsilon/3$-differentially private.

		\item To compute $\tilde{\mu}_{\bg}^v$: 
		In Equation \eqref{eq:AVG_query_B}, for a voltage level $v$, the mean value $\mu_{\bg}^v$ of the conductance vector $\bg$ the computation of is computed issuing an average query $Q_A$ on $\bg$. The privacy budget used is $\frac{n_v\,\epsilon}{3\alpha}$. 
		Therefore, by Property \ref{prop:sens_avg} and Theorem \ref{th:m_lap} computing $\tilde{\mu}_{\bg}^v$ is $\epsilon/3$-differentially private. Computing the vector of mean values $\tilde{\mu}_\bg = \{\tilde{\mu}_{\bg}^{v_1}, \ldots, \tilde{\mu}_{\bg}^{v_{|VL(\cN)|}}\}$ for each voltage level of the network is also $\epsilon/3$-differentially private by parallel composition (Theorem \ref{th:par_composition}) since each line belongs to exactly one voltage level set $E(v)$.
	
		\item To compute $\tilde{\mu}_{\bb}^v$: 
		The mean value $\mu_{\bb}$ is $\epsilon/3$ differentially private. The argument is analogous to the one above. 
	\end{enumerate}
	Computing a privacy-preserving version of the susceptance vector $\bb$ uses exclusively privacy-preserving data ($\bg$) and public information ($\bm r$), therefore the vector $\tilde{\bb}$ defined in Equation \ref{eq:ID_query} is differentially private by post-processing immunity (Theorem \ref{th:postprocessing}). 
	
	Note that the optimization model of line (5) uses \emph{exclusively} privacy-preserving data $(\tilde{\bg}, \tilde{\bb}, \tilde{\mu}_\bg, \tilde{\mu}_\bb)$ and additional public information (i.e., the optimization problem and its optimal solution value).
	The result follows by sequential composition (Theorem \ref{th:seq_composition}) and post-processing immunity (Theorem \ref{th:postprocessing}).
\end{proof}

\begin{customthm}{5}
	For a given $\alpha$-indistinguishability level, the MPLO mechanism is $\epsilon$-differential private.
\end{customthm}

\begin{proof}
	Note that MPLO differs from PLO exclusively in the optimization model executed on line 4. The optimization model does not operate on the sensitive data and takes as input the same privacy-preserving values as those taken as input by the PLO's optimization model. Therefore, by post-process immunity and Theorem \ref{thm4} the MPLO mechanism is $\epsilon$-differential private.
\end{proof}

\section{MPLO Algorithm}
\label{sec:MPLO-alg}

The MPLO mechanism is outlined in Algorithm \ref{alg:mplo}.
It takes as input a sequence of power network descriptions $\{ \cN_t \}_{t=1}^h$ where the loads and the optimal generator dispatches are varying over time, the associated optimal dispatch costs $\{O^*(t)\}_{t=1}^h$, as well as three positive real numbers: $\epsilon$, which determines the \emph{privacy value} of the private data, $\alpha$, which determines the \emph{indistinguishability value}, and $\beta$, which determines the required \emph{faithfulness} to the value of the objective function. 
Recall that each $\cN_t = (N, E, \bS_t, \bY, \bm{\theta^\Delta}, \bs, \bv)$ represents the steady state of the power grid at time step $1 \leq t \leq h$, and therefore $\{ \cN_t \}_{t=1}^h$ represents a network of \emph{time-series} data.

Lines 1 to 4 of Algorithm \ref{alg:mplo} are analogous to lines 1 to 4 of Algorithm \ref{alg:plo} executed by the PLO mechanism. 
They produce the noisy conductance and susceptance vectors (lines 1 and 2) $\tilde{\bg}$ and $\tilde{\bb}$, respectively, as well as their noisy mean values (lines 3 and 4). Since the voltage levels $VL(\cN_t)$ are invariant for any $t=1,\ldots,h$, the loop of line 3 uses network $\cN_1$ as a reference. 
Line 5 describe the post-processing optimization step associated to the mechanism, and, similarly to that associated to the PLO mechanism, it operates using exclusively the privacy-preserving version of the line parameters.

The multi-step PLO post-processing finds the set of conductance $\dot{\bg}$ and susceptance $\dot{\bb}$ values, summarized with the notation $\dot{\bY}$, by minimizing Equation \eqref{obj:mplo}. This process is similar to the one performed by the (single-step) PLO post-processing (in Equation \eqref{dp_obj}). The models differ in their constraints. While the PLO post-processing operates over a single network, the MPLO post-processing works over $h$ networks. 
Constraints \eqref{c:m2} and \eqref{c:m3} are similar to Constraint \eqref{dp_bound_g} and \eqref{dp_bound_b}, respectively, of Algorithm \ref{alg:plo}. They bound the values for the post-processed conductance and susceptance to be close to their privacy-preserving means, for each voltage level $v$ and parametrized by a value $\lambda > 0$. These constraints are used to avoid that the post-processed values deviate arbitrarily from the original ones.  
Constraint \eqref{c:m4} guarantees to satisfy condition
\ref{tag:cond3} of the PLO problem (Definition \ref{def:plop}) for each network $\cN_t$ in the horizon ($t \in [h]$). The notation $\bm{cost}(S^g_i)$ is used as a shorthand for $c_{2i} (\Re(S^g_i))^2 + \bm
c_{1i}\Re(S^g_i) + \bm c_{0i}$. 
Finally, Constraints \eqref{c:m5} to \eqref{c:m11} enforce the AC-OPF Constraints \eqref{eq:ac_0} to \eqref{dp_ac_4} for each network $\cN_t$ in the considered horizon $(t \in [h])$/

\section{Extended Results}

\begin{figure*}[!t]
	\centering
	\includegraphics[width=.75\linewidth]{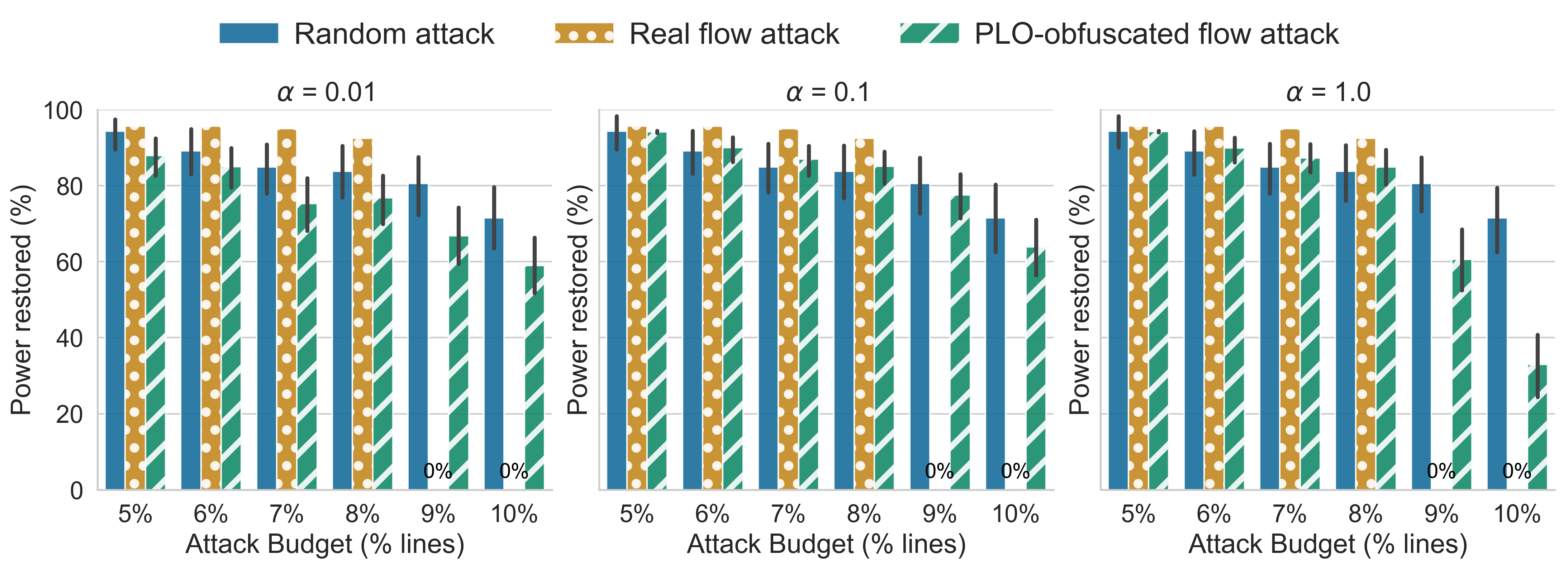}

	\caption{Percentage of the active load restored after different attacks on the IEEE-118 bus network with $\alpha=0.01$ (left), $0.1$ (center), and $1.0$ (right), and $\beta=0.01$.}
	\label{fig:attack_3-a}
\end{figure*}

\begin{algorithm}[!t]
    \setcounter{AlgoLine}{0}
	\caption{The MPLO mechanism for the AC-OPF}
	\label{alg:mplo}
	\SetKwInOut{Input}{input}
    \SetKwInOut{Output}{output}	
	\Input{$\langle \{\cN_t\}_{t=1}^h, 
				    \{\cO^*(t)\}_{t=1}^h, \epsilon, \alpha, \beta \rangle$}

	$\tilde{\bg} \gets \bg + Lap(\frac{3\alpha}{\epsilon})$ \\
	$\tilde{\bb} \gets \textbf{r} \cdot \tilde{\bg} $ \\
	\ForEach{$v \in \textsl{VL}(\cN_1)$}{

	$\tilde{\mu}_{\bx}^v \gets 
		\frac{1}{n} \sum_{(ij)\in E(v)} x_{ij} 
		+ Lap(\frac{3\alpha}{n_v\epsilon})$ 
		(for $x = {\bg, \bb}$)
	}
	Solve the following model:
	\nonl
	{\small
	\begin{align}
	\mbox{\bf variables: }
	& S^g_i (t), V_i (t)      \hspace*{75pt} \forall i \in N, t \in [h] \nonumber \\
	& \dot{Y}_{ij}, S_{ij}(t) \hspace*{36pt}  \forall (i,j)\in E_k \!\cup\! E_k^R, \forall t \in [h] \nonumber\\
	\mbox{\bf minimize: }  
	&  \| \dot{\bg} - \tilde{\bg} \|_2^2  + \| \dot{\bb} - \tilde{\bb} \|_2^2   \label{obj:mplo} \tag{$m_1$} \\
	\mbox{\bf subject to:} & \notag \\
	& \hspace*{-30pt} 
		\forall v \in \textsl{VL}(\cN_1),\; \forall (i,j) \in E_k(v) \cup E_k^R(v): \notag\\ 
	& \hspace*{-20pt} 
		\quad \frac{1}{\lambda}\mu_{\bg}^v \leq \dot{g}_{ij} \leq \lambda\,\mu_{\bg}^v \label{c:m2} \tag{$m_{2}$}\\
	& \hspace*{-20pt} 
		\quad \frac{1}{\lambda}\mu_{\bb}^v \leq \dot{b}_{ij} \leq \lambda\,\mu_{\bb}^v \label{c:m3} \tag{$m_{3}$}\\ 
	& \hspace*{-30pt} 
	\forall t \in [h]: \notag\\
    & \hspace*{-20pt} 
    	\frac{\big|\sum_{i \in \cN_t} \bm{cost} (S^g_i(t)) - \cO^*(t) \big|}{\cO^*(t)} \leq \beta \label{c:m4}  \tag{$m_4$}\\
	& \hspace*{-20pt} 
		\angle V_{s}(t) = 0 \label{c:m5} \tag{$m_5$} \\
 	& \hspace*{-20pt} 
	 	\forall i \in \cN_t: & \nonumber \\
	& \hspace*{-20pt} 
		\quad \bm {v^l}_i \leq |V_i(t)| \leq \bm {v^u}_i  \label{c:m6} \tag{$m_6$}\\
	& \hspace*{-20pt} 
		\quad \bm {S^{gl}}_i \leq S^g_i(t) \leq \bm {S^{gu}}_i \label{c:m7}  \tag{$m_7$}\\	    
	& \hspace*{-20pt} 
		\quad S^g_i(t) - {\bm S^d_i(t)} = \textstyle\sum_{(i,j)\in E_k \cup E_k^R} S_{ij}(t) \label{c:m8}  \tag{$m_8$}\\ 
	& \hspace*{-20pt} 
		\forall (i,j) \in E_k \cup E_k^R: \nonumber \\
	& \hspace*{-20pt} 
		\quad -\bm {\theta^\Delta}_{ij} \leq \angle (V_i(t) V^*_j(t)) \leq \bm {\theta^\Delta}_{ij} \label{c:m9}  \tag{$m_9$}\\	    
	& \hspace*{-20pt} 
		\quad |S_{ij}(t)| \leq \bm {s^u}_{ij}           \label{c:m10}  \tag{$m_{10}$}\\
	& \hspace*{-20pt} 
		\quad S_{ij}(t) = \dot{Y}^*_{ij} |V_i(t)|^2 - \dot{Y}^*_{ij} V_i(t) V^*_j(t) \label{c:m11} \tag{$m_{11}$}
     \end{align}\\
	}
	\Output{$\{\dot{\cN}_t\}_{t=1}^h = 
		\langle  N, E, \bS_t, \dot{\bY}, \bm{\theta^\Delta}, \bs, \bv \rangle$}
\end{algorithm}

This section presents additional results to shed additional lights on the attack performance of the proposed PLO mechanism on the IEEE-118 bus network. To do so, Figure~\ref{fig:attack_3-a} illustrates the percentage of the load being restored for the three introduced attack strategies at a finer granularity of the attack budget $k$ from 5\% to 10\%, at varying of the indistinguishability value $\alpha \in \{0.01, 0.1, 1.0\}$ and with faithfulness value $\beta = 0.01$.  
The results report the average values of 100 simulations for each combination of parameters.

The plots show that the \emph{real flow attacks} require to damage as little as $9\%$ of the network power lines to produce un-restorable damages. 
We observe that the network restoration abilities after a real flow attack decrease drastically when the attack budget increases from $8\%$ to $9\%$. This is due to that there exist a set of critical power lines, that, if collectively damaged results in an un-restorable network. On the other hand, when only a subset of these power lines is damaged, a high percentage of the network loads can be restored. 

On the other hand, the PLO mechanism is able to hide the critical power lines to an attacker exploiting the released data, thus avoiding such critical restoration behavior. 

\begin{figure}[!h]
	\centering
	\includegraphics[width=\linewidth]{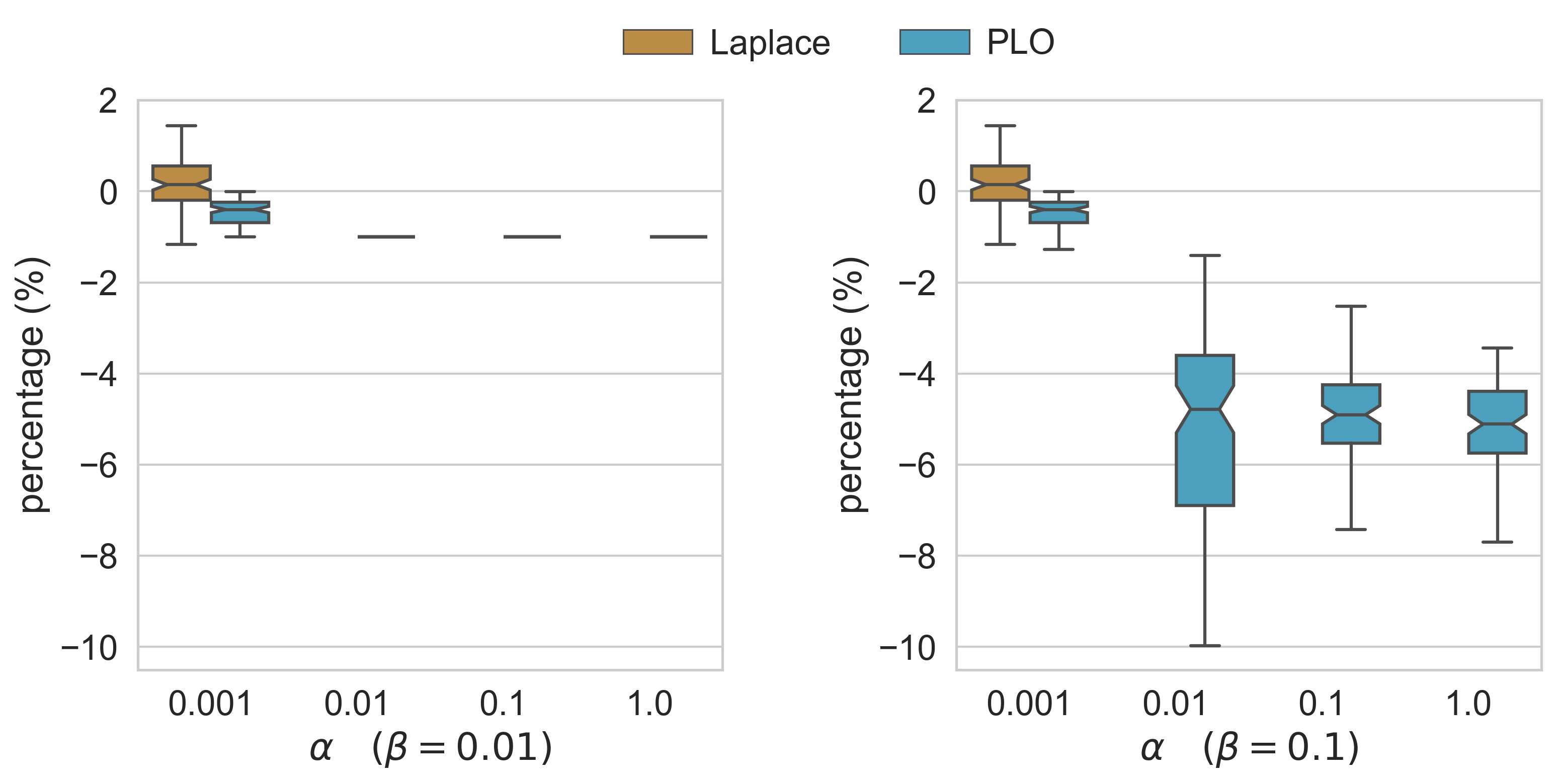}
	\caption{The IEEE-39 bus AC optimal dispatch costs differences, in percentage, between the original and the obfuscated networks with faithfulness parameters $\beta=0.01$ (left) and $0.1$ (right).}
	\label{fig:10}
\end{figure}
 \begin{figure}[!h]
	\centering
	\includegraphics[width=\linewidth]{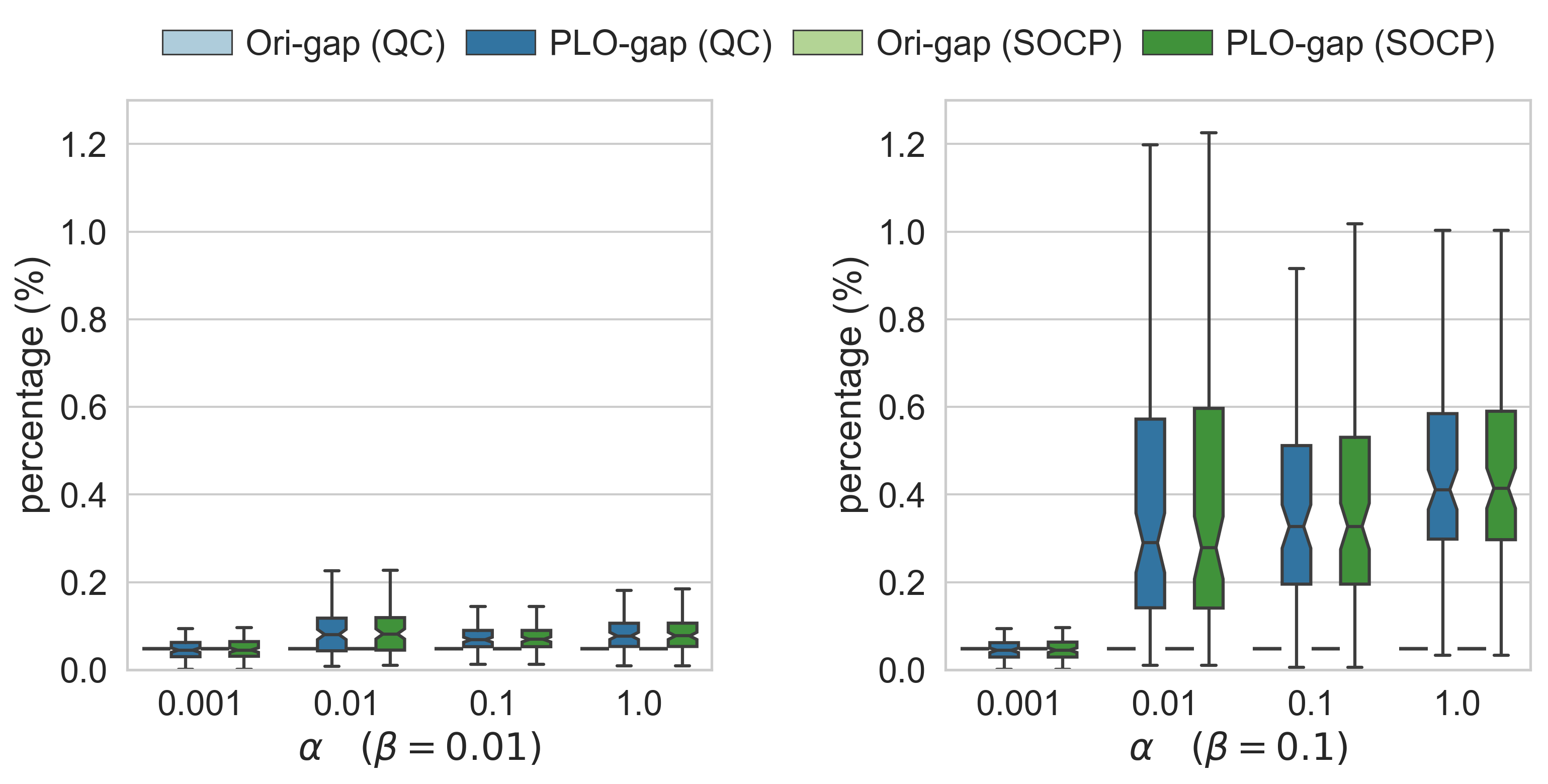}
	\caption{The IEEE-39 bus optimality gap differences, in percentage, between the original and the obfuscated networks with faithfulness parameters $\beta=0.01$ (left) and $0.1$ (right).}
	\label{fig:11}
\end{figure}
Figures \ref{fig:10} and \ref{fig:11} report the boxplot
visualizations of Figures \ref{fig:opf_12a} and \ref{fig:opf_12b},
respectively.

\end{document}